\documentclass{article} 
\usepackage{iclr2027_conference,times}

\usepackage{amsmath,amsfonts,bm}

\def\eqref#1{Equation~\ref{#1}}

\def\1{\bm{1}}

\def\vtheta{{\bm{\theta}}}
\def\vbeta{{\bm{\beta}}}
\def\vxi{{\bm{\xi}}}
\def\vvarepsilon{{\bm{\varepsilon}}}

\def\vb{{\bm{b}}}
\def\vc{{\bm{c}}}
\def\vd{{\bm{d}}}

\def\vg{{\bm{g}}}

\def\vw{{\bm{w}}}
\def\vx{{\bm{x}}}
\def\vy{{\bm{y}}}

\def\mB{{\bm{B}}}
\def\mC{{\bm{C}}}

\def\mE{{\bm{E}}}

\def\mH{{\bm{H}}}

\def\mL{{\bm{L}}}

\def\mQ{{\bm{Q}}}

\def\mU{{\bm{U}}}
\def\mV{{\bm{V}}}

\def\mX{{\bm{X}}}

\def\mLambda{{\bm{\Lambda}}}
\def\mSigma{{\bm{\Sigma}}}
\def\mDelta{{\bm{\Delta}}}

\DeclareMathAlphabet{\mathsfit}{\encodingdefault}{\sfdefault}{m}{sl}
\SetMathAlphabet{\mathsfit}{bold}{\encodingdefault}{\sfdefault}{bx}{n}

\newtheorem{theorem}{\bf{Theorem}}
\newtheorem{lemma}{\bf{Lemma}}

\newenvironment{theorem*}[1][]{%
    \par\addvspace{\topsep}
    \noindent{\bfseries Theorem\ifx&#1&\relax\else\ (#1)\fi\quad}\itshape
    \ignorespaces
}{%
    \par\addvspace{\topsep}
}

\newenvironment{lemma*}[1][]{%
    \par\addvspace{\topsep}
    \noindent{\bfseries Lemma\ifx&#1&\relax\else\ (#1)\fi\quad}\itshape
    \ignorespaces
}{%
    \par\addvspace{\topsep}
}

\newenvironment{proposition*}[1][]{%
    \par\addvspace{\topsep}
    \noindent{\bfseries Proposition\ifx&#1&\relax\else\ (#1)\fi\quad}\itshape
    \ignorespaces
}{%
    \par\addvspace{\topsep}
}

\newenvironment{corollary*}[1][]{%
    \par\addvspace{\topsep}
    \noindent{\bfseries Corollary\ifx&#1&\relax\else\ (#1)\fi\quad}\itshape
    \ignorespaces
    }{%
    \par\addvspace{\topsep}
}

\newenvironment{remark*}[1][]{%
    \par\addvspace{\topsep}
    \noindent{\bfseries Remark\ifx&#1&\relax\else\ (#1)\fi\quad}\itshape
    \ignorespaces
}{%
    \par\addvspace{\topsep}
}

\usepackage{graphicx}
\usepackage{hyperref}
\usepackage{url}
\usepackage{booktabs}
\usepackage{multirow}
\usepackage{float}
\usepackage{xcolor}

\definecolor{bestborder}{RGB}{55,145,105}
\definecolor{bestfill}{RGB}{225,245,235}
\definecolor{secondborder}{RGB}{85,125,190}
\definecolor{secondfill}{RGB}{230,238,250}

\newcommand{\se}[1]{\text{\scriptsize\,(#1)}}

\title{Adaptive Transfer of Representational \\ Signatures for LLM Reliability Detection}

\author{Yongtong Li \\
Northwestern University \\
\texttt{yongtongli2026@u.northwestern.edu}
\And
Yixuan Florence Wu \\
Northwestern University \\
\texttt{yixuanwu2029@u.northwestern.edu}
\And
Jiawei Zhang \\
University of Wisconsin--Madison \\
\texttt{jzhang2924@wisc.edu}
\And
Naichen Shi \\
Northwestern University \\
\texttt{naichen.shi@northwestern.edu}
}

\iclrfinalcopy 
\begin{document}

\maketitle
\pagestyle{plain}
\begin{abstract}
Large language models (LLMs) are reported to be susceptible to reliability issues including hallucinations and reasoning errors. Existing quality detectors can predict unreliable outputs based on internal representations with decent accuracies. However, training these detectors often requires substantial domain-specific labeled data, which limits their applicability to new domains. We propose to address this challenge by transferring representational signatures across related domains. More specifically, we build a statistical framework that first extracts low-dimensional signatures for each domain, then uses cross-fitting to learn composition weights of domain-specific predictors. The weights adapt to each source domain’s predictive relevance to the target domain. Our asymptotic analysis of a least-squares analogue characterizes the limiting prediction risk and aggregation weights in terms of domain discrepancies and representation bias. Experiments across multiple hallucination datasets and open-weight LLMs show that, with only 25 labeled questions in the new domain, adaptive transfer improves hallucination-detection AUROC by 6.8\%. For reasoning-step prediction, it improves AUROC by 8.4\%.
\end{abstract}

\section{Introduction}
Large language models (LLMs) have demonstrated strong performance across a wide range of tasks, including language generation \citep{brown2020language}, mathematical reasoning \citep{lewkowycz2022solving}, and code generation \citep{chen2021codex}. However, strong performance on these tasks does not necessarily imply reliable behavior. Hallucinations and misalignment can lead to incorrect decisions \citep{kaissis2024clinical,hu2026understanding} or unintended actions \citep{zhang2025breakingagents}. In July 2026, for example, OpenAI reported that models undergoing internal cybersecurity evaluations with reduced safeguards bypassed containment controls and compromised Hugging Face's production infrastructure \citep{openai2026huggingface}. These risks highlight the need to detect unreliable LLM behaviors.

Recent studies have linked unreliable behavior in open-weight LLMs to specific patterns in their internal representations. For hallucination detection, for example, linear probes trained on hidden-state activations from a single layer can distinguish correct answers from hallucinated ones with high accuracy \citep{oneill2025single, orgad2025llmsknow}. Other approaches reduce the dimensionality of internal representations before classification \citep{hu2026harp}, or capture more complex patterns using nonlinear classifiers, such as multilayer perceptrons (MLPs) \citep{azaria2023internal, zhou2025hademif} and vision transformers (ViTs) \citep{barshalom2025actvit}. Together, these methods suggest that internal representations provide informative signals for hallucination detection.

Despite their success, many of these methods rely on labeled examples from a specific dataset or domain~\citep{shamsi2026density}. In practice, labeled examples can be costly to obtain in sufficient quantities for every new domain. Existing work addresses this by reshaping internal states to make truthfulness signals more consistent across domains \citep{zhang2025prism} or by constructing hallucination scores from uncertainty fluctuations that remain stable under domain shift \citep{deng2026spikescore}.  However, transfer performance can vary substantially across domains. A detector trained on a single dataset may generalize better to some related domains while still performing poorly on others  \citep{alansari2026crosshallu}. This raises a natural question:

\textit{Can we adaptively recompose reliability detectors trained on multiple related source domains to achieve accurate detection in a target domain with limited labeled data?}

To answer this question, we develop an adaptive transfer method that learns a target-domain detector using information from multiple source domains. For each source domain, we train a classifier on a low-dimensional principal component subspace that captures the dominant variation in the model's internal representations from perturbed prompts. We then linearly combine these source detectors with a detector trained on the available labeled target-domain data. Because the source detectors may differ in their predictive relevance to the target domain, we use cross-fitting to learn composition weights that balance information transferred from the source domains with the information learned directly from the target domain. We call the resulting method Adaptive Signature Transfer (AdaST).

To understand when source-domain information benefits AdaST, we develop an asymptotic analysis of the adaptive transfer procedure in the regression setting. The analysis characterizes how the learned composition weights converge to the population-optimal counterparts, which are dependent on the discrepancy between source and target domains. The theoretical analysis confirms that AdaST asymptotically performs better than or as well as the predictor that only relies on the target domain information.

We use two tasks to validate the numerical performance of AdaST, including hallucination detection and mathematics reasoning correctness prediction. AdaST achieves robust predictive performance across multiple domains on multiple open-weight models. Furthermore, we show that the learned detector vectors can be used as steering directions. Our main contributions are summarized as:
\begin{enumerate}
    \item We develop an adaptive cross-domain transfer method, AdaST, for representational signatures of LLM behavior. The method constructs principal-component detectors in each domain and uses cross-fitting to learn adaptive transfer weights. 
    \item We characterize the limiting behavior of the adaptive transfer mechanism in the regression setting, which validates the effectiveness of adaptive transfer when the target domain is similar to the source domains. 
    \item We demonstrate empirically that internal-state detectors can benefit from cross-domain transfer, with larger gains when fewer labeled examples are available in the target domain.
    
\end{enumerate}

\subsection{Related Work}

\paragraph{Representation-based Reliability Detection.} Existing representation-based LLM reliability detection methods have explored a variety of designs, including linear probes applied directly to model activations \citep{oneill2025single, orgad2025llmsknow} and low-dimensional subspace methods that extract compact behavioral representations \citep{hu2026harp}. More flexible neural classifiers have also been investigated, such as MLPs trained on hidden state activations \citep{azaria2023internal,zhou2025hademif} and Vision Transformers that model the structured layer-by-token organization of activation tensors \citep{barshalom2025actvit}. These methods can achieve strong performance when sufficient labeled data are available, but their effectiveness can degrade under label scarcity \citep{shamsi2026density}.

\paragraph{Cross-Domain Reliability Detection.}

Recent work examines whether reliability signals learned from internal representations can directly transfer across domains. \citet{orgad2025llmsknow} report substantial cross-dataset degradation, while \citet{alansari2026crosshallu} find that transfer can succeed but varies with the training and test datasets. Other work improves cross-domain robustness by modifying internal representations or constructing more stable detection signals \citep{zhang2025prism,hu2026harp,deng2026spikescore}. These methods train a detector on a single source domain and apply it unchanged to a new domain.  Our method instead studies how detectors learned from multiple source domains can be adaptively combined for a new domain with limited labeled data.

\paragraph{Multi-Source Transfer Learning.} 

Multi-source transfer learning studies how knowledge from several related source domains can be combined to improve prediction in a target domain. Early studies formalized the multi-source adaptation problem and developed principled ways to combine source-specific predictors \citep{mansour2008domain,mansour2009multiple}. Subsequent methods have explored a variety of strategies for source combination, including mixture-of-experts models with target-dependent weights \citep{guo2018moe}, representation alignment \citep{peng2019moment}, and stacking or similarity-based weighting \citep{wu2023regression}. We build on these multi-source transfer learning ideas to study adaptive transfer across representation-based LLM reliability detectors.

\section{Method}
\label{sec:method}

\begin{figure}[H]
    \centering
    \includegraphics[width=\linewidth]{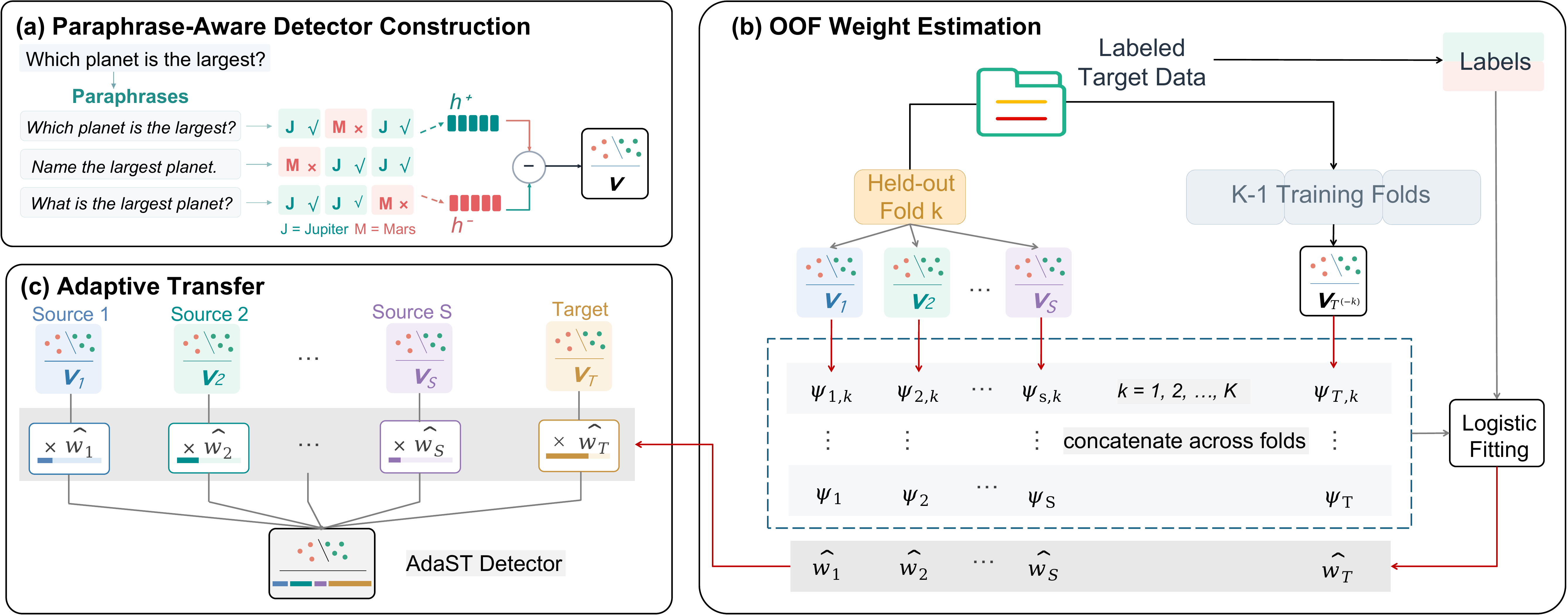}
    \caption{Overview of the proposed AdaST framework. }
    \label{fig:schematic}
\end{figure}

In this section, we present the proposed AdaST framework for cross-domain binary behavioral prediction, as illustrated in Figure~\ref{fig:schematic}. First, for each domain, we query the LLM with multiple semantically equivalent paraphrases of the same questions and extract the final-token hidden states as representations of the generated responses. We then construct question-level contrast vectors by comparing hidden states associated with the two behavioral classes and use these contrasts to estimate a low-dimensional behavioral subspace and train a domain-specific detector. Second, the learned detectors are adaptively combined using weights estimated from cross-fitted predictions in the target domain. 

\subsection{Paraphrase-Based Behavioral Subspace}

Let $[S]=\{1,\ldots,S\}$ index the source domains and $T$ index the target domain. For each domain index $s\in[S]\cup\{T\}$, let $\mathcal{Q}_s$ denote the set of questions from the corresponding domain. For example, $\mathcal{Q}_1$ may contain factual question-answering prompts such as ``What is the capital of France?'', while $\mathcal{Q}_2$ may contain conversational question-answering prompts such as ``Which city do you recommend visiting in France?'' The semantic similarity across these domains allows us to transfer knowledge about LLM representations.

Fix a domain index $s\in[S]\cup\{T\}$. For each question $q\in\mathcal{Q}_s$, we retain the original wording of $q$ and generate $P-1$ semantically equivalent paraphrases, inspired by~\citet{song2026inv}. For example, the question ``What is the capital of France?'' can be paraphrased as ``Name the capital of France.'' We then query the LLM $R$ times for each version of the question $q$ to obtain $R$ responses per version. For each generated response, we extract the hidden state at the final generated token from a selected model layer and denote it by $\mathbf h_{qpr}\in\mathbb{R}^{D}$, where $D$ is the hidden-state dimension, $p$ indexes the question version, and $r$ indexes the generated response. Each response has a binary behavioral label $y_{qpr}\in\{0,1\}$, where $y_{qpr}=1$ denotes the positive class and $y_{qpr}=0$ the negative class. Let $\mathcal Q_s^{\pm}\subseteq\mathcal Q_s$ denote the questions for which both behavioral classes are observed, and let $m_s=|\mathcal Q_s^{\pm}|$.

Following prior work on contrastive representation learning~\citep{zou2023representation,li2023inference,rimsky2024steering}, we represent the behavioral difference for each question by comparing the mean representations associated with its positive and negative responses. Let $\mathcal I_q^+=\{(p,r):y_{qpr}=1\}$ and
$\mathcal I_q^-=\{(p,r):y_{qpr}=0\}$
denote the positive and negative classes for question $q$, respectively. For each $q\in\mathcal Q_s^{\pm}$, we define the contrast vector:
\begin{equation}
\boldsymbol{\delta}_q
=
\frac{1}{|\mathcal I_q^{+}|}
\sum_{(p,r)\in\mathcal I_q^{+}} \mathbf h_{qpr}
-
\frac{1}{|\mathcal I_q^{-}|}
\sum_{(p,r)\in\mathcal I_q^{-}} \mathbf h_{qpr}.
\label{eq:question-difference}
\end{equation}

Here, $\boldsymbol{\delta}_q$ quantifies the representation difference between the positive and negative behavioral classes for the same question $q$. We stack the contrast vectors as rows of $\mathbf{C}_s$, define their mean as $\bar{\boldsymbol{\delta}}_s
=\frac{1}{m_s}\sum_{q\in\mathcal Q_s^{\pm}}\boldsymbol{\delta}_q$, and center them across questions to obtain $\mathbf{C}_s-\mathbf{1}\bar{\boldsymbol{\delta}}_s^\top$. Since $m_s$ may be small relative to the hidden-state dimension $D$, the problem becomes overparameterized. Consequently, fitting a detector directly in the full residual space can be unstable and prone to overfitting. We therefore perform principal component analysis on the centered contrast matrix to obtain a lower-dimensional representation:
\begin{equation}
\mC_s-\mathbf{1}\bar{\boldsymbol{\delta}}_s^\top
=
\mL_s\mLambda_s\mV_s^\top,
\qquad
\mB_s=\mV_{s,1:k_s}^{\top}\in\mathbb{R}^{k_s\times D},
\label{eq:domain-basis}
\end{equation}
We retain $k_s=\min\{k_{\max},m_s-1,D\}$ principal directions, where $k_{\max}$ is selected using held-out validation data, and the rows of $\mB_s$ span the domain-specific behavioral subspace. 

For a hidden state $\mathbf h\in\mathbb{R}^D$, we first project it onto the behavioral subspace $\mB_s$. Let $\boldsymbol{\mu}_s$ and
$\boldsymbol{\gamma}_s$ denote the coordinate-wise mean and standard deviation
of $\mB_s\mathbf h$ over the training responses in domain $s$. For numerical stability, we standardize the projected representations coordinate-wise and define 
\begin{equation}
\boldsymbol{\phi}_s(\mathbf h)
=
\operatorname{diag}(\boldsymbol{\gamma}_s)^{-1}
(\mB_s\mathbf h-\boldsymbol{\mu}_s).
\label{eq:standardized-projection}
\end{equation}

We then fit a logistic regression in this standardized subspace and denote its coefficient vector by $\boldsymbol{\beta}_s$. The same procedure is applied independently to each source domain and the target domain.

\subsection{Adaptive Cross-Domain Transfer}
\label{sec:adaptive-transfer}
When labeled data in a target domain are scarce, a detector trained only on the target domain may generalize poorly to unseen examples. Meanwhile, labeled datasets from related domains are often available. When the underlying tasks are similar, the corresponding internal representations, here taken as residual-stream activations, may share structure across domains. We therefore supplement the target detector with detectors learned from the source domains. 

The domain-specific detectors are learned in different low-dimensional coordinate systems induced by their respective bases $\mB_s$. Since these coordinate systems are not directly comparable, we map each domain-specific coefficient vector $\boldsymbol{\beta}_s$ back to the shared residual space. Using the standardized projection in \eqref{eq:standardized-projection}, the coefficient on $\mathbf h$ is 
\begin{equation}
\mathbf v_s
=
\mB_s^\top\operatorname{diag}(\boldsymbol{\gamma}_s)^{-1}\boldsymbol{\beta}_s.
\label{eq:residual-direction}
\end{equation}
Since the detector scores are standardized over the labeled target responses before combination, the offset has no effect on transfer. Only the direction of $\mathbf v_s$ matters after this standardization, so we normalize it as $\mathbf v_s \leftarrow \mathbf v_s/\|\mathbf v_s\|_2$ and use $\mathbf v_s^\top\mathbf h$ to score target hidden states.

The target detector is fitted using the labeled target data, so its predictions on the same examples are in-sample. Using these predictions to learn the combination weights can therefore favor the target detector over the source detectors. To remove such an advantage, we use a $K$-fold cross-fitting scheme. Specifically, let $n_T$ denote the number of labeled target questions available for training. These questions are partitioned into $K$ equal folds. For each held-out fold, the target subspace and detector are refitted on the other $K-1$ folds, and the corresponding residual-space direction $\mathbf v_T$ is recomputed, while the source detectors remain fixed. For each held-out fold $k$, let $\operatorname{Std}_{-k}(\cdot)$ denote standardization using the other $K-1$ target folds. We apply $\operatorname{Std}_{-k}$ to the detector scores $\mathbf v_s^\top\mathbf h$ on the held-out fold and concatenate the standardized scores over all $K$ folds to form the cross-fitted score vector $\boldsymbol{\psi}_s$. These score vectors are concatenated as

\[
\mathbf{\Psi}
=
\begin{bmatrix}
\boldsymbol{\psi}_1 &
\cdots &
\boldsymbol{\psi}_S &
\boldsymbol{\psi}_T
\end{bmatrix}.
\]

Let $\mathbf w=(w_1,\ldots,w_S,w_T)^\top$ denote the combination weights, where $w_1,\ldots,w_S$ correspond to the source detectors and $w_T$ to the target detector, and let $b$ denote the intercept. Let $i$ index the labeled target responses $(q,p,r)$. For each response $i$, let $y_i$ be its binary label and $\mathbf{\Psi}_i$  the corresponding row of $\mathbf{\Psi}$. We estimate the combination weights and intercept once from the complete $\mathbf{\Psi}$ by solving the constrained logistic regression problem:
\begin{equation}
    (\widehat b,\widehat{\mathbf w})
    =
    \arg\min_{\substack{b\in\mathbb R,\; \mathbf w\geq0\\
    \mathbf{1}^\top\mathbf w=1}}
    \sum_i
    \left[
    \log\!\left(1+\exp(b+\mathbf{\Psi}_i\mathbf w)\right)
    -y_i(b+\mathbf{\Psi}_i\mathbf w)
    \right].
    \label{eq:simplex-mixture}
\end{equation}

\eqref{eq:simplex-mixture} learns the transfer allocation weights from the cross-fitted scores. The learned weights determine how much each detector contributes to the target prediction, while the intercept adjusts the overall offset of the combined score. For prediction, we train the target detector on all labeled target questions so that all available target supervision is used for prediction. For the fixed source detectors and the trained target detector, we compute the score standardization statistics using all labeled target responses and denote the standardized score for a hidden state $\mathbf h$ by $\psi_s(\mathbf h)$. For a new target response, we combine the standardized scores from the fixed source detectors and the trained target detector using the learned weights. This defines the AdaST detector:

\begin{equation}
\widehat f(\mathbf h)
=
\widehat b
+
\sum_{s\in[S]\cup\{T\}}
\widehat w_s \psi_s(\mathbf h).
\end{equation}

$\widehat f(\mathbf h)$ is a scalar logit for the new target response. Applying the sigmoid function to $\widehat f(\mathbf h)$ gives the predicted probability that the response belongs to the positive class. Larger values therefore indicate stronger evidence for the positive class, whereas smaller values indicate stronger evidence for the negative class.

\section{Asymptotic Analysis}
\label{sec:asymptotic-analysis}

\subsection{Risk Analysis}
To characterize the geometry of transfer, we study a least-squares analogue of the detector in Section~\ref{sec:method}, using fixed prediction subspaces and convex combinations of unstandardized linear predictions. This formulation characterizes the limiting weights and prediction risk in terms of domain discrepancy and representation bias.

For each domain $s\in[S]\cup\{T\}$ with $n_s$ samples, consider the linear model $y_s=\vx_s^\top\vtheta_s^\star+\varepsilon_s$, where $\vx_s\sim\mathcal N(0,\mSigma_s)$ and $\varepsilon_s$ is independent noise with mean zero and variance $\sigma_s^2$. Let the orthonormal columns of $\mU_s\in\mathbb R^{D\times k}$ span a fixed prediction subspace, which corresponds to $\mB_s^\top$ in Section~\ref{sec:method}. The population projected coefficient and the matrix of prediction discrepancies relative to the target are
\[
\vtheta_s^\circ
=\mU_s(\mU_s^\top\mSigma_s\mU_s)^{-1}\mU_s^\top\mSigma_s\vtheta_s^\star,
\qquad
\mQ=\big[\vtheta_1^\circ-\vtheta_T^\star,\ldots,
\vtheta_S^\circ-\vtheta_T^\star,\vb_T\big],
\]
where $\vb_T=\vtheta_T^\circ-\vtheta_T^\star$ is the target representation bias. We work in the asymptotic limit of source and target sample sizes, while holding $S$, $D$ and $k$ fixed. Assume that
$\overline{\mH}:=\mQ^\top\mSigma_T\mQ\succ0$. We estimate weights by minimizing two-fold target out-of-fold squared loss subject to $\vw\ge0$ and $\mathbf1^\top\vw=1$, then refit the target predictor on all target observations. The last coordinate of the weight vector corresponds to the target domain, with $w_{S+1}=w_T$. $\widehat{\vtheta}_{S+1}$ denotes the target-only coefficient refitted on all target observations. Write $\widehat{\vtheta}_{\text{AdaST}}$ for the aggregate and $\mathcal R_T$ for its squared prediction risk on an independent target observation. The full estimator and regularity assumptions are given in Appendix~\ref{app:proof-risk}.

\begin{theorem}[Asymptotic risk and weight convergence]
\label{thrm:risk}
Under the assumptions in Appendix~\ref{app:risk-assumptions},
\begin{align*}
\widehat{\vw}
&\xrightarrow{a.s.}\vw^\star
=\arg\min_{\substack{\vw\ge0\\\mathbf1^\top\vw=1}}
\vw^\top\overline{\mH}\vw,\\
\mathcal R_T(\widehat{\vtheta}_{\text{AdaST}})
&\xrightarrow{a.s.}
\sigma_T^2+(\vw^\star)^\top\overline{\mH}\vw^\star
\le\sigma_T^2+\overline H_{S+1,S+1}.
\end{align*}
The limiting weights are unique. Equality with the target-only risk holds exactly when $\vw^\star=\mathbf e_{S+1}$, the target-only weight vector.

Moreover, if $n_s\ge c n_T$ for all $s\in[S]$ and some fixed $c>1$, then $ \mathbb E|\widehat w_T-w_T^\star|=O(n_T^{-1/2})$. 
\end{theorem}

The proof and finite-sample bounds are given in Appendices~\ref{app:proof-risk} and~\ref{app:finite-rates}. The asymptotic result shows how complementary source predictors can reduce target representation bias, yielding a limiting risk no greater than that of the target-only predictor. Under the stated sample-size conditions, the finite-sample rate quantifies how quickly the learned target weight approaches its population-optimal value and gives a vanishing bound on its expectation when sources alone are optimal.

\subsection{Synthetic Validation}
\label{sec:synthetic- validation}
We study nonnegative least-squares aggregation with coefficient shifts $\mDelta_s=\vtheta_s^\star-\vtheta_T^\star=\tau\vd_s$, where $\vd_s\in\operatorname{col}(\mU_s)$ and $\frac1S\sum_s\vd_s^\top\mSigma_T\vd_s=1$. The projected discrepancies satisfy $\widetilde\mDelta_s(\tau)=\vtheta_s^\circ(\tau)-\vtheta_T^\star=\vc_s+\tau\vd_s$, with $\vc_s=\widetilde\mDelta_s(0)$. $\tau$ quantifies how far the source coefficients drift from the target coefficient, with smaller values corresponding to source coefficients closer to the target coefficient, while larger values indicate greater source--target heterogeneity. Even at $\tau=0$, different prediction subspaces can produce different projected coefficients. We consider $\tau\in\{0.4,0.5,0.6,0.75\}$ and $n_T\in\{100,200,400,800,1600,3200\}$, with $200$ replications per setting.

\begin{figure}[ht]
    \centering
    \IfFileExists{figures/transfer_mass_panels_nonnegative.png}{%
        \includegraphics[width=0.98\linewidth]{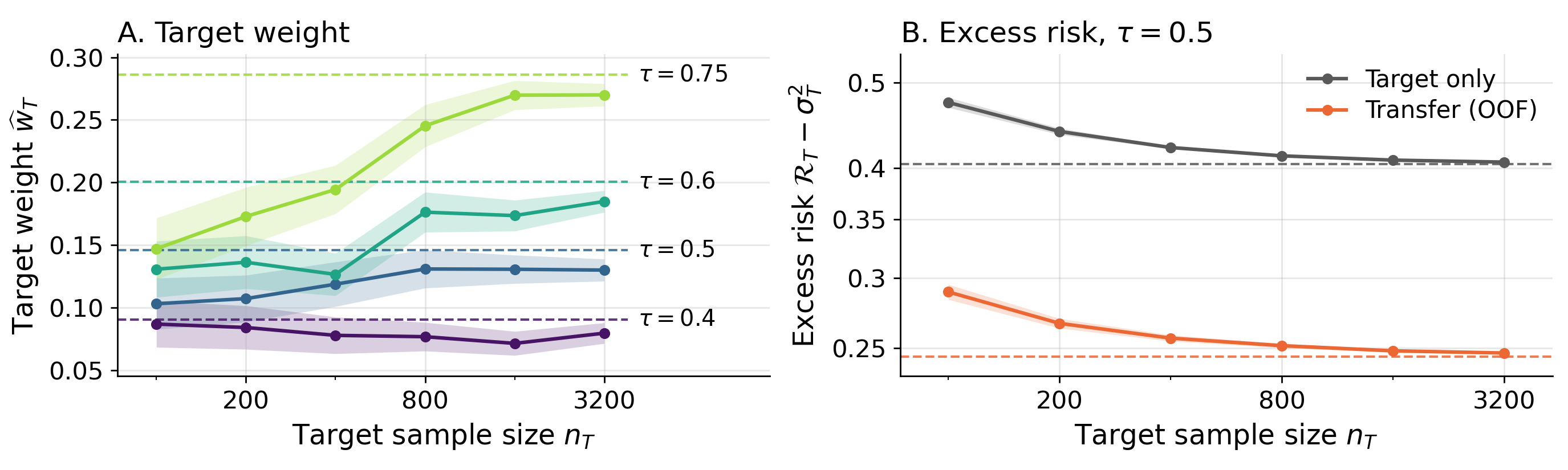}%
    }{%
        \fbox{\parbox{0.9\linewidth}{\centering
        Synthetic validation with nonnegative weights is awaiting rerun.}}%
    }
    \caption{Synthetic validation of simplex-constrained least-squares aggregation. \textbf{A.} Target weight $\widehat w_T$ against $n_T$ for several source drifts $\tau$. \textbf{B.} Excess risk $\mathcal R_T-\sigma_T^2$ against $n_T$ at $\tau=0.5$. Solid lines summarize Monte Carlo means with $95\%$ confidence bands for the means. Dashed lines mark the constrained population weights and risks from Theorem~\ref{thrm:risk}.}
    \label{fig:sim}
\end{figure}

Figure~\ref{fig:sim}~\textbf{A} shows that for larger $\tau$, the learned target weight $\widehat w_T$ approaches its limit $w_T^\star(\tau)$ as the sample sizes grow, and this limit increases with $\tau$. Intuitively, larger values of $\tau$ imply that the source domains are less informative for the target, so more weight shifts to the target. In this regime, the empirical target weight generally increases with $n_T$, as additional target observations improve the target fit and reduce the need to borrow information from mismatched source domains. Figure~\ref{fig:sim}~\textbf{B} shows that the excess risks $\mathcal R_T-\sigma_T^2$ of the transfer and target-only estimators converge to their limits, with the transfer limit strictly below the target-only one.

\section{Experiments}
\label{sec:experiments}
In this section, we evaluate AdaST primarily on cross-domain hallucination detection. We further examine its generalization to reasoning-step correctness. We also analyze the learned representations and evaluate activation steering based on the resulting directions. 

\subsection{Experimental Setup}
\label{sec:experimental-setup}

\paragraph{Tasks and datasets.}
We consider several open-weight language models, including OPT-6.7B \citep{zhang2022opt}, Llama-3.1-8B \citep{grattafiori2024llama}, Gemma-7B \citep{gemmateam2024gemma}, Mistral-7B \citep{jiang2023mistral}, and DeepSeek-LLM-7B \citep{deepseekai2024deepseek}.  Across these models, we evaluate two behavioral prediction tasks, hallucination detection and reasoning-step correctness classification. Hallucination detection is evaluated on Natural Questions \citep{kwiatkowski2019natural}, TriviaQA \citep{joshi2017triviaqa},
CoQA \citep{reddy2019coqa}, HaluEval-QA \citep{li2023halueval},
SimpleQA \citep{wei2024simpleqa}, and HotpotQA \citep{yang2018hotpotqa}. 
Reasoning-step correctness is evaluated on PRM800K \citep{lightman2023lets}, which contains questions from seven mathematical domains, including Algebra, Counting and Probability, Geometry, Intermediate Algebra, Number Theory, Prealgebra, and Precalculus. 

\paragraph{Evaluation metrics.}
We use AUROC as the primary discrimination metric, defined as $\mathrm{AUROC}=\Pr(s(x^+)>s(x^-))$. Here, $s(\cdot)$ denotes the detector score and $x^+$ and $x^-$ denote positive and negative responses. We also report Macro-F1, defined as $\mathrm{Macro\text{-}F1}=\frac{1}{2}\left(F1_{\mathrm{positive}}+F1_{\mathrm{negative}}\right)$, where $F1_c=\frac{2P_cR_c}{P_c+R_c}$ for a class $c$. We report Macro-F1 rather than the F1 score of either class because the response labels can be imbalanced. Macro-F1 gives equal weight to truthful and hallucinated responses and therefore avoids a high score driven primarily by the majority class.

\paragraph{Baselines.}
We compare against five hallucination detection baselines. HARP~\citep{hu2026harp} derives a reasoning subspace from the model's unembedding matrix and performs detection on hidden states projected onto this subspace. SinkProbe~\citep{binkowski2026sinkprobe} detects hallucinations using statistics of attention-sink patterns derived from attention maps. HaloScope~\citep{du2024haloscope} uses a latent subspace to separate unlabeled generations into candidate truthful and hallucinated sets, which are then used to train a binary truthfulness  classifier. PRISM-SAPLMA~\citep{zhang2025prism} applies PRISM prompts before extracting hidden states and feeds the resulting representations to SAPLMA for cross-domain detection. CrossHallu~\citep{alansari2026crosshallu} trains an MLP detector on source-domain features derived from hidden states, attention, and token probabilities, and studies its transfer to unseen target domains.

\paragraph{Implementation details.}

For hallucination detection, we use six QA datasets as six domains. Each dataset is treated in turn as the target domain, with the remaining five datasets serving as source domains. Each question is augmented with two semantically equivalent paraphrases, giving three question versions in total. We generate five responses per version and label each response by comparing it with all available reference answers using BLEURT-20 ~\citep{sellam2020bleurt,pu2021learning}. A response is labeled positive if its maximum BLEURT score exceeds $0.5$ and negative otherwise. The target-label budget $n_T$ counts original questions, with all paraphrases and generated responses from the same question kept together.

For reasoning-step correctness, we use PRM800K reasoning prefixes from the seven MATH subjects. In our transfer setting, each subject is treated in turn as the target, with the remaining six serving as sources. For each prefix, the evaluated model generates eight candidate next steps, which are scored by Qwen2.5-Math-PRM-7B~\citep{zhang2025lessons}. We retain the highest-scoring candidate with score at least $0.8$ as positive and the lowest-scoring candidate with score at most $0.2$ as negative, discarding prefixes for which both classes are unavailable. The target-label budget $n_T$ counts retained reasoning prefixes.

For both tasks, representations and domain-specific detectors are constructed as described in Section~\ref{sec:method}, and transfer weights are estimated using three-fold cross-fitting. Model-specific layer selection are provided in Appendix~\ref{app:layer}.

\subsection{Main Results}
\label{sec:main-results}

\begin{table}[ht]
\centering
\caption{Performance of target-only detectors and AdaST across different target-label budgets, averaged over six datasets for hallucination detection and seven PRM800K MATH domains for reasoning-step correctness. Subscripts report standard errors.}
\label{tab:ours-two-tasks}
\small
\setlength{\tabcolsep}{4pt}
\resizebox{\textwidth}{!}{%
\begin{tabular}{ll*{4}{cc}}
\toprule
& & \multicolumn{2}{c}{$n_T=25$} & \multicolumn{2}{c}{$n_T=50$} & \multicolumn{2}{c}{$n_T=100$} & \multicolumn{2}{c}{Full} \\
\cmidrule(lr){3-4} \cmidrule(lr){5-6} \cmidrule(lr){7-8} \cmidrule(lr){9-10}
Task & Data access & AUC & F1 & AUC & F1 & AUC & F1 & AUC & F1 \\
\midrule
Hallucination detection & Target-only
& 74.2\se{1.0} & 62.8\se{1.5} & 77.3\se{0.9} & 65.1\se{1.4}
& 79.8\se{0.8} & 66.9\se{1.3} & 84.9\se{0.5} & 71.5\se{1.1} \\
 & AdaST
& \textbf{81.0}\se{0.9} & \textbf{66.2}\se{1.4} & \textbf{81.7}\se{0.8} & \textbf{67.3}\se{1.4}
& \textbf{82.3}\se{0.8} & \textbf{68.4}\se{1.3} & \textbf{85.0}\se{0.5} & \textbf{71.5}\se{1.1} \\
\cmidrule(l){2-10}
Reasoning-step correctness & Target-only
& 81.1\se{1.0} & 71.9\se{1.0} & 85.5\se{1.0} & 76.3\se{0.8}
& 87.6\se{0.9} & 78.8\se{1.0} & 88.9\se{0.9} & 80.7\se{0.9} \\
 & AdaST
& \textbf{89.5}\se{0.9} & \textbf{80.8}\se{1.0} & \textbf{89.8}\se{0.9} & \textbf{81.2}\se{1.0}
& \textbf{90.2}\se{0.9} & \textbf{82.2}\se{1.1} & \textbf{90.1}\se{0.8} & \textbf{82.2}\se{0.9} \\
\bottomrule
\end{tabular}}
\end{table}

Table~\ref{tab:ours-two-tasks} compares the target-only detector with AdaST across the two tasks and different target-label budgets.  AdaST consistently improves both AUROC and Macro-F1 across the two tasks, with larger gains when fewer target labels are available. With limited target data, the target detector is estimated with greater uncertainty. AdaST can compensate for this by assigning weight to source detectors that remain informative on held-out target examples. As more target labels become available, the target detector becomes more reliable and the additional benefit of source information decreases. Results by model and target dataset are reported in Appendices~\ref{app:per-model} and~\ref{app:per-dataset}.

\begin{figure}[ht]
    \centering
    \includegraphics[width=0.9\linewidth]{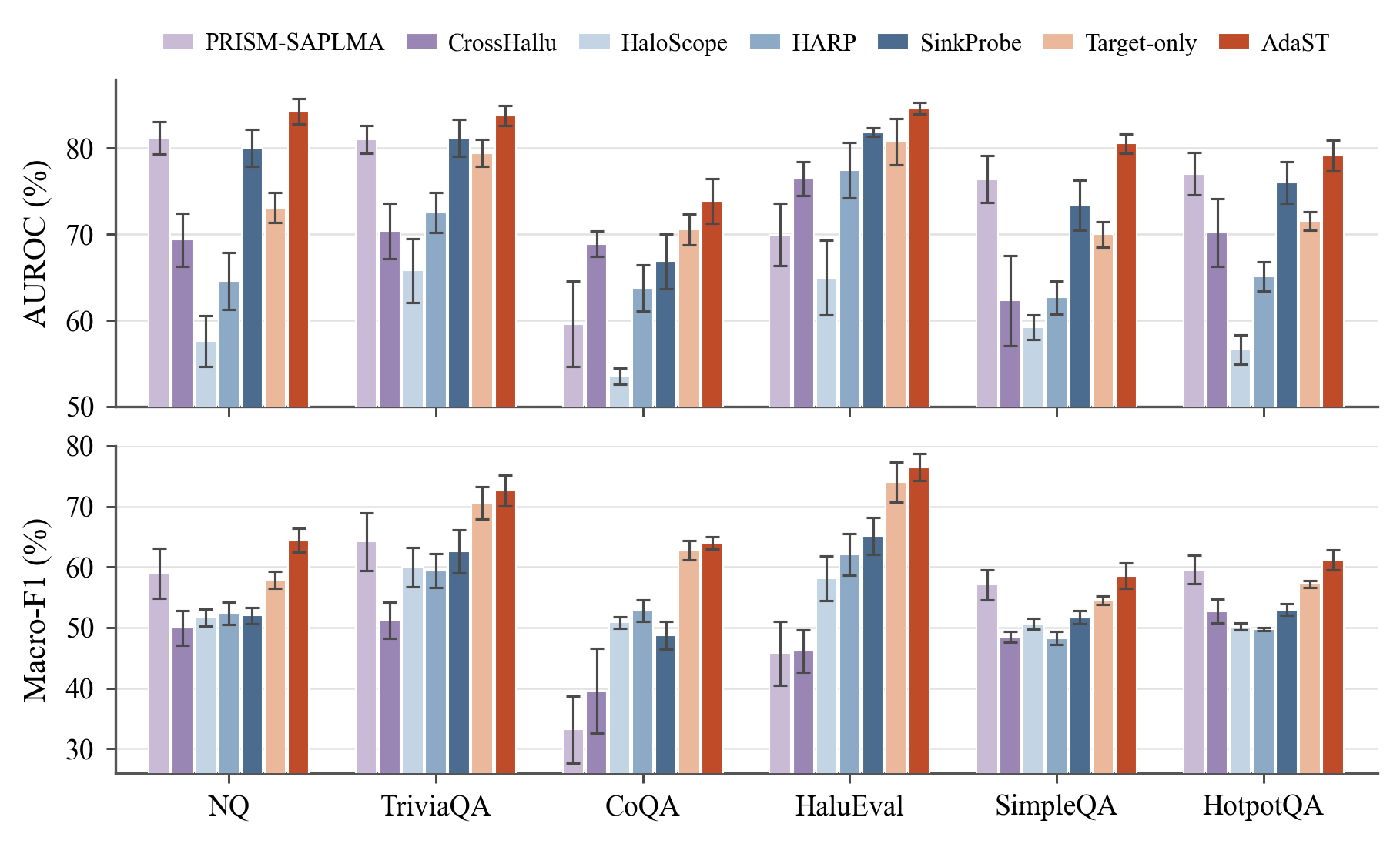}
    \caption{Hallucination detection at $n_T=25$, by target dataset. Each bar is the mean over the five evaluated models and error bars are standard errors across models. }
    \label{fig:baseline-comparison}
\end{figure}

Figure~\ref{fig:baseline-comparison} compares AdaST with existing hallucination detectors at $n_T=25$. AdaST improves over the domain-specific detector across all models and achieves the strongest average performance in both AUROC and Macro-F1. We also observe that some baselines remain competitive in AUROC but perform worse in Macro-F1. This gap shows that strong ranking performance does not always lead to balanced classification. AdaST remains strong on both metrics. Results by model are reported in Appendices~\ref{app:baselines}.

\subsection{Further Analysis}
\label{sec:further-analysis}

\begin{figure}[ht]
    \centering
    \includegraphics[width=\linewidth]{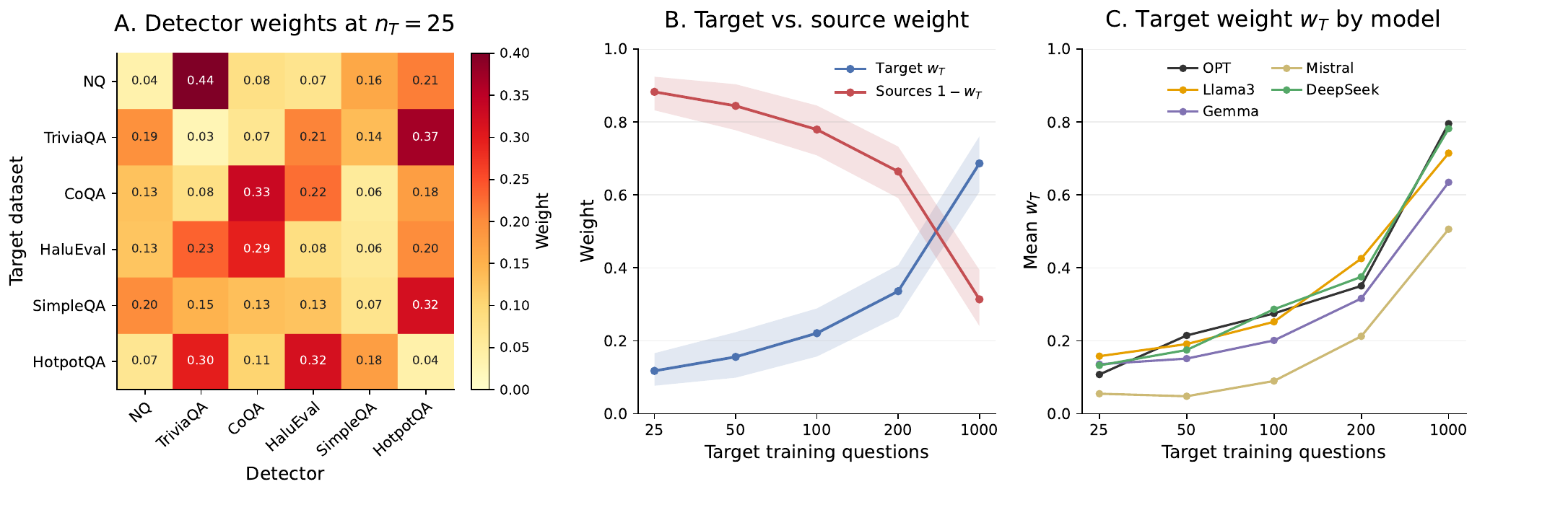}
    \caption{Learned AdaST combination weights. (a) Detector weights at $n_T=25$. (b) Target weight $w_T$ and source weight $1-w_T$ across target training sizes. (c) Target weight $w_T$ across model families.
}
    \label{fig:weight-path}
\end{figure}

\paragraph{Learned transfer path.} Figure~\ref{fig:weight-path}~\textbf{A} shows the learned expert allocation at $n_T=25$. The weights differ across target datasets, suggesting that different targets benefit from different source detectors rather than from a common source combination. Figure~\ref{fig:weight-path}~\textbf{B} shows how the overall allocation changes with target supervision. With few target questions, most weights are placed on the source detectors. As $n_T$ grows, the target detector becomes more reliable, while the relative benefit of source information diminishes, leading AdaST to place more weight on the target detector. Figure~\ref{fig:weight-path}~\textbf{C} shows the same overall trend across model families.

\paragraph{Robustness to base detector choice.}

To test whether AdaST is robust to the choice of base detector, we use SinkProbe, the strongest single-dataset baseline in Figure~\ref{fig:baseline-comparison}. For each domain, a SinkProbe expert is a logistic classifier fitted to its attention-sink features. The resulting classifier score is used as the domain-specific detector score in AdaST. Table~\ref{tab:sinkprobe-transfer} reports the results.

\begin{table}[ht]
\caption{Robustness to base detector choice at $n_T=25$. AUROC (\%).}
\label{tab:sinkprobe-transfer}
\centering
\small
\begin{tabular}{lcccccc}
\toprule
SinkProbe as base detector & OPT-6.7B & Mistral-7B & Llama-3.1-8B & DeepSeek-7B & Gemma-7B\\
\midrule
Target-only & $71.8$\se{4.6} & $81.4$\se{2.4} & $79.4$\se{3.5} & $72.9$\se{3.4} & $80.4$\se{3.8}\\
AdaST & $\mathbf{73.3}$\se{3.9} & $\mathbf{82.7}$\se{3.4} & $\mathbf{81.7}$\se{3.5} & $\mathbf{74.6}$\se{3.7} & $\mathbf{81.8}$\se{4.2}\\
\bottomrule
\end{tabular}
\end{table}

Table~\ref{tab:sinkprobe-transfer} shows that AdaST improves SinkProbe across five models, with an average gain of 1.61 AUROC points. SinkProbe relies on attention-sink features rather than the residual-stream representations used by our default detector. This suggests that the transfer gains are not specific to a particular detector construction.

\paragraph{Effect of paraphrase-aware estimation.} Table~\ref{tab:paraphrase-ablation} examines the effect of paraphrase augmentation. Within the evaluated range, adding paraphrased views improves AUROC for both the target-only detector and AdaST. The improvement  suggests that paraphrases help produce a more stable domain-specific behavioral
representation.
\begin{table}[ht]
\centering
\caption{Effect of paraphrase augmentation (AUROC, \%).}
\label{tab:paraphrase-ablation}
\small
\setlength{\tabcolsep}{9pt}
\begin{tabular}{lccc}
\toprule
Method & $P=1$ & $P=2$ & $P=3$ \\
\midrule
Target-only        & 69.6\se{4.0} & 71.9\se{2.9} & \textbf{72.8}\se{2.5} \\
AdaST & 76.2\se{2.8} & 77.7\se{2.3} & \textbf{78.2}\se{2.3} \\
\bottomrule
\end{tabular}
\end{table}

\paragraph{From Detection to Intervention.}
We test whether the directions learned for detection can also steer generation. Using the residual-space directions in \eqref{eq:residual-direction} and the transfer weights in \eqref{eq:simplex-mixture}, we define
\[
\mathbf u_{\mathrm{mix}}
=
\frac{
\sum_{s\in[S]\cup\{T\}} \widehat w_s\,\mathbf v_s/\kappa_s
}{
\left\|
\sum_{s\in[S]\cup\{T\}} \widehat w_s\,\mathbf v_s/\kappa_s
\right\|_2
},
\]
where $\kappa_s$ is the target-score normalization factor for detector $s$.
During generation, we steer the residual state at the detection layer as $\mathbf h_t^{(L)} \leftarrow \mathbf h_t^{(L)}+\alpha\mathbf u_{\mathrm{mix}}, $ where $\alpha>0$ controls the steering strength. The computation of $\kappa_s$ and the full activation-steering implementation are detailed in Appendix~\ref{app:intervention}.

\begin{table}[ht]
\centering
\small
\caption{Activation steering results on Llama-3.1-8B, averaged over three seeds. }
\begin{tabular}{lccc}
\toprule
Setting & Semantic accuracy & Hallucination & Abstention  \\
\midrule
No steering      & 27.2 & 67.8 & 5.0 \\
AdaST steering  & \textbf{30.0}\se{0.7} & \textbf{64.6}\se{0.9} & 5.4\se{0.4} \\
\bottomrule
\end{tabular}
\label{tab:steering}
\end{table}

We use Phi-4 \citep{abdin2024phi4} to assign semantic labels, from which we report semantic accuracy, hallucination and abstention. Abstention is the fraction of responses that explicitly decline to answer or provide no substantive answer. Steering improves semantic accuracy and reduces hallucination. The abstention rate remains largely unchanged, suggesting that the improvement is not simply driven by the model avoiding difficult questions.

\section{Conclusion}

We study adaptive transfer of LLM reliability signals across domains with AdaST. Experiments on hallucination detection and reasoning-step correctness show that transfer is most beneficial when target labels are limited, with the learned weights gradually shifting toward the target detector as more supervision becomes available. Our analysis characterizes this behavior in terms of source--target discrepancies and target representation bias. Future research will include extending linear classifiers to nonlinear ones.

\section*{AI Use Statement}
Generative AI tools were used to assist with language editing, figure presentation, and feedback on the manuscript. All AI-assisted content was reviewed and verified by the authors, who take full responsibility for the final content.

\bibliography{iclr2027_conference}
\bibliographystyle{iclr2027_conference}

\appendix

\section{Additional Results}

\subsection{Comparisons Across Models}
\label{app:per-model}

Table~\ref{tab:ours-per-model} breaks the $n_T=25$ column of Table~\ref{tab:ours-two-tasks} down by model. The gain from the source domains is positive for every model on both tasks.

\begin{table}[H]
\centering
\caption{Per-model breakdown of Table~\ref{tab:ours-two-tasks} at $n_T=25$.}
\label{tab:ours-per-model}
\small
\setlength{\tabcolsep}{3pt}
\resizebox{\textwidth}{!}{%
\begin{tabular}{ll*{6}{cc}}
\toprule
& & \multicolumn{2}{c}{OPT-6.7B}
& \multicolumn{2}{c}{Llama-3.1-8B}
& \multicolumn{2}{c}{Gemma-7B}
& \multicolumn{2}{c}{Mistral-7B}
& \multicolumn{2}{c}{DeepSeek-7B}
& \multicolumn{2}{c}{Avg.} \\
\cmidrule(lr){3-4}
\cmidrule(lr){5-6}
\cmidrule(lr){7-8}
\cmidrule(lr){9-10}
\cmidrule(lr){11-12}
\cmidrule(lr){13-14}
Task & Data access & AUC & F1 & AUC & F1 & AUC & F1 & AUC & F1 & AUC & F1 & AUC & F1 \\
\midrule
Hallucination & Target-only & 74.1\se{2.3} & 61.4\se{3.5} & 74.1\se{2.7} & 65.2\se{3.7} & 74.9\se{2.7} & 64.4\se{4.0} & 72.4\se{1.1} & 59.3\se{1.0} & 75.7\se{2.0} & 63.9\se{3.1} & 74.2\se{1.0} & 62.8\se{1.5} \\
 & AdaST & \textbf{78.6}\se{2.2} & \textbf{62.9}\se{2.9} & \textbf{82.6}\se{1.8} & \textbf{69.7}\se{2.9} & \textbf{81.4}\se{2.1} & \textbf{66.5}\se{3.7} & \textbf{83.0}\se{0.5} & \textbf{66.3}\se{1.0} & \textbf{79.6}\se{2.3} & \textbf{65.6}\se{3.2} & \textbf{81.0}\se{0.9} & \textbf{66.2}\se{1.4} \\
\cmidrule(l){2-14}

Reasoning-step & Target-only & 84.8\se{1.5} & 74.7\se{1.5} & 80.2\se{1.9} & 71.9\se{2.5} & 73.3\se{1.8} & 65.1\se{1.6} & 83.5\se{1.2} & 73.7\se{1.4} & 83.7\se{1.2} & 74.0\se{1.3} & 81.1\se{1.0} & 71.9\se{1.0} \\
 & AdaST & \textbf{92.3}\se{0.8} & \textbf{84.8}\se{0.6} & \textbf{87.5}\se{1.3} & \textbf{78.3}\se{1.3} & \textbf{82.5}\se{1.9} & \textbf{72.7}\se{1.9} & \textbf{92.4}\se{0.5} & \textbf{83.4}\se{0.8} & \textbf{92.8}\se{1.0} & \textbf{84.9}\se{1.5} & \textbf{89.5}\se{0.9} & \textbf{80.8}\se{1.0} \\
\bottomrule
\end{tabular}}
\end{table}

\label{app:reasoning-base}

\subsection{Comparisons Across Dataset}
\label{app:per-dataset}

Tables~\ref{tab:hallu-budget} and~\ref{tab:reasoning-budget} break Table~\ref{tab:ours-two-tasks} down by target dataset.  The gain from the source domains is positive on every target dataset of both tasks at $n_T=25$.

\begin{table}[H]
\centering
\caption{AdaST for hallucination detection against the same detector without source data, by dataset and target label budget $n_T$. }
\label{tab:hallu-budget}
\small
\setlength{\tabcolsep}{3pt}
\resizebox{\textwidth}{!}{%
\begin{tabular}{ll*{7}{cc}}
\toprule
& & \multicolumn{2}{c}{NQ} & \multicolumn{2}{c}{TriviaQA} & \multicolumn{2}{c}{CoQA} & \multicolumn{2}{c}{HaluEval} & \multicolumn{2}{c}{SimpleQA} & \multicolumn{2}{c}{HotpotQA} & \multicolumn{2}{c}{Avg.} \\
\cmidrule(lr){3-4} \cmidrule(lr){5-6} \cmidrule(lr){7-8} \cmidrule(lr){9-10} \cmidrule(lr){11-12} \cmidrule(lr){13-14} \cmidrule(lr){15-16}
$n_T$ & Data access & AUC & F1 & AUC & F1 & AUC & F1 & AUC & F1 & AUC & F1 & AUC & F1 & AUC & F1 \\
\midrule
$25$ & Target-only & 73.1\se{1.5} & 57.9\se{1.3} & 79.5\se{1.4} & 70.6\se{2.4} & 70.6\se{1.6} & 62.8\se{1.4} & 80.8\se{2.4} & 74.0\se{3.0} & 70.0\se{1.3} & 54.6\se{0.6} & 71.5\se{1.0} & 57.2\se{0.5} & 74.2\se{1.0} & 62.8\se{1.5} \\
 & AdaST & \textbf{84.3}\se{1.3} & \textbf{64.4}\se{1.7} & \textbf{83.8}\se{1.0} & \textbf{72.6}\se{2.3} & \textbf{73.8}\se{2.3} & \textbf{64.0}\se{0.9} & \textbf{84.6}\se{0.6} & \textbf{76.5}\se{2.0} & \textbf{80.5}\se{1.0} & \textbf{58.6}\se{1.9} & \textbf{79.2}\se{1.6} & \textbf{61.2}\se{1.5} & \textbf{81.0}\se{0.9} & \textbf{66.2}\se{1.4} \\
\cmidrule(l){2-16}
$50$ & Target-only & 77.4\se{2.0} & 60.9\se{1.6} & 80.4\se{1.4} & 71.3\se{2.5} & 74.1\se{1.9} & 65.8\se{1.1} & 83.0\se{1.6} & 75.5\se{2.8} & 73.2\se{0.6} & 56.5\se{1.0} & 75.9\se{1.3} & 60.3\se{0.6} & 77.3\se{0.9} & 65.1\se{1.4} \\
 & AdaST & \textbf{84.7}\se{1.4} & \textbf{65.4}\se{1.9} & \textbf{84.2}\se{0.9} & \textbf{74.0}\se{1.9} & \textbf{75.4}\se{2.3} & \textbf{66.3}\se{1.0} & \textbf{85.1}\se{0.6} & \textbf{77.1}\se{2.1} & \textbf{81.0}\se{1.0} & \textbf{59.5}\se{1.7} & \textbf{79.7}\se{1.6} & \textbf{61.4}\se{1.8} & \textbf{81.7}\se{0.8} & \textbf{67.3}\se{1.4} \\
\cmidrule(l){2-16}
$100$ & Target-only & 81.1\se{1.6} & 63.4\se{1.7} & 82.3\se{1.2} & 72.8\se{2.2} & 75.9\se{2.0} & 67.0\se{1.5} & 84.4\se{1.0} & 76.8\se{2.4} & 77.8\se{0.7} & 59.1\se{0.8} & 77.6\se{1.2} & 62.1\se{0.9} & 79.8\se{0.8} & 66.9\se{1.3} \\
 & AdaST & \textbf{85.4}\se{1.4} & \textbf{66.6}\se{1.5} & \textbf{84.3}\se{0.9} & \textbf{74.7}\se{1.9} & \textbf{76.8}\se{2.0} & \textbf{67.9}\se{1.0} & \textbf{85.3}\se{0.6} & \textbf{77.6}\se{2.1} & \textbf{81.3}\se{1.0} & \textbf{60.4}\se{1.7} & \textbf{80.7}\se{1.3} & \textbf{63.2}\se{1.1} & \textbf{82.3}\se{0.8} & \textbf{68.4}\se{1.3} \\
\cmidrule(l){2-16}
Full & Target-only & 86.4\se{1.3} & 68.2\se{1.3} & 86.0\se{0.8} & 76.4\se{1.7} & \textbf{81.6}\se{1.3} & \textbf{72.7}\se{1.1} & 87.6\se{0.9} & 78.9\se{2.1} & 84.0\se{0.8} & 64.7\se{1.1} & 84.2\se{0.8} & \textbf{67.8}\se{0.4} & 85.0\se{0.5} & 71.5\se{1.1} \\
 & AdaST & \textbf{86.6}\se{1.3} & \textbf{68.5}\se{1.5} & \textbf{86.1}\se{0.7} & \textbf{76.5}\se{1.7} & 81.6\se{1.3} & 72.7\se{1.2} & \textbf{87.6}\se{0.9} & \textbf{78.9}\se{2.1} & \textbf{84.1}\se{0.7} & \textbf{64.8}\se{1.1} & \textbf{84.2}\se{0.8} & 67.7\se{0.4} & \textbf{85.0}\se{0.5} & \textbf{71.5}\se{1.1} \\
\bottomrule
\end{tabular}}
\end{table}

\begin{table}[H]
\centering
\caption{AdaST for Reasoning-step correctness against the same detector without source data, by MATH domain and target label budget $n_T$.  Prealgebra has fewer than $100$ labeled training questions, so the $n_T=100$ row omits it and averages over the remaining six domains.}
\label{tab:reasoning-budget}
\small
\setlength{\tabcolsep}{3pt}
\resizebox{\textwidth}{!}{%
\begin{tabular}{ll*{8}{cc}}
\toprule
& & \multicolumn{2}{c}{Algebra} & \multicolumn{2}{c}{C.\&Prob.} & \multicolumn{2}{c}{Geometry} & \multicolumn{2}{c}{Int.\ Alg.} & \multicolumn{2}{c}{N.\ Theory} & \multicolumn{2}{c}{Prealg.} & \multicolumn{2}{c}{Precalc.} & \multicolumn{2}{c}{Avg.} \\
\cmidrule(lr){3-4} \cmidrule(lr){5-6} \cmidrule(lr){7-8} \cmidrule(lr){9-10} \cmidrule(lr){11-12} \cmidrule(lr){13-14} \cmidrule(lr){15-16} \cmidrule(lr){17-18}
$n_T$ & Data access & AUC & F1 & AUC & F1 & AUC & F1 & AUC & F1 & AUC & F1 & AUC & F1 & AUC & F1 & AUC & F1 \\
\midrule
$25$ & Target-only & 77.1\se{3.5} & 67.6\se{4.1} & 78.8\se{2.5} & 70.6\se{2.3} & 82.6\se{1.0} & 73.9\se{0.4} & 82.9\se{2.4} & 73.8\se{2.5} & 81.1\se{2.3} & 70.5\se{1.8} & 82.3\se{2.9} & 74.0\se{2.2} & 82.8\se{1.9} & 72.7\se{1.7} & 81.1\se{1.0} & 71.9\se{1.0} \\
 & AdaST & \textbf{85.6}\se{3.7} & \textbf{77.3}\se{3.9} & \textbf{89.2}\se{1.5} & \textbf{81.0}\se{1.6} & \textbf{91.3}\se{1.2} & \textbf{81.9}\se{1.9} & \textbf{90.4}\se{2.0} & \textbf{81.1}\se{2.8} & \textbf{90.9}\se{1.6} & \textbf{82.0}\se{2.6} & \textbf{87.8}\se{2.2} & \textbf{79.8}\se{2.3} & \textbf{91.4}\se{1.3} & \textbf{82.7}\se{1.3} & \textbf{89.5}\se{0.9} & \textbf{80.8}\se{1.0} \\
\cmidrule(l){2-18}
$50$ & Target-only & 81.9\se{3.6} & 73.9\se{3.2} & 85.1\se{2.1} & 77.0\se{1.6} & 87.5\se{1.4} & 76.5\se{1.4} & 85.7\se{2.7} & 77.4\se{2.6} & 86.7\se{2.1} & 77.6\se{2.1} & 84.0\se{2.5} & 74.0\se{1.7} & 87.6\se{1.8} & 78.0\se{1.8} & 85.5\se{1.0} & 76.3\se{0.8} \\
 & AdaST & \textbf{86.1}\se{3.8} & \textbf{78.2}\se{3.9} & \textbf{89.3}\se{1.3} & \textbf{80.1}\se{1.5} & \textbf{91.5}\se{1.4} & \textbf{83.0}\se{2.6} & \textbf{90.9}\se{1.9} & \textbf{81.4}\se{2.6} & \textbf{90.9}\se{1.6} & \textbf{83.5}\se{1.7} & \textbf{88.3}\se{2.4} & \textbf{80.1}\se{2.7} & \textbf{91.6}\se{1.4} & \textbf{81.9}\se{1.7} & \textbf{89.8}\se{0.9} & \textbf{81.2}\se{1.0} \\
\cmidrule(l){2-18}
$100$ & Target-only & 84.0\se{3.3} & 75.6\se{3.3} & 87.7\se{1.6} & 77.9\se{1.8} & 88.1\se{1.8} & 78.1\se{2.1} & 88.0\se{1.8} & 79.5\se{2.5} & 88.1\se{2.1} & 80.1\se{2.2} & -- & -- & 89.9\se{1.5} & 81.7\se{1.3} & 87.6\se{0.9} & 78.8\se{1.0} \\
 & AdaST & \textbf{86.5}\se{3.5} & \textbf{78.9}\se{3.5} & \textbf{89.4}\se{1.4} & \textbf{81.6}\se{1.7} & \textbf{91.5}\se{1.5} & \textbf{83.7}\se{2.5} & \textbf{91.0}\se{1.8} & \textbf{82.6}\se{2.4} & \textbf{91.0}\se{1.6} & \textbf{83.9}\se{2.3} & -- & -- & \textbf{91.8}\se{1.3} & \textbf{82.1}\se{1.9} & \textbf{90.2}\se{0.9} & \textbf{82.2}\se{1.1} \\
\cmidrule(l){2-18}
Full & Target-only & 85.9\se{3.1} & 77.5\se{3.1} & 89.0\se{1.5} & 81.3\se{1.3} & 89.8\se{1.8} & 82.8\se{2.2} & 90.9\se{1.8} & 83.0\se{2.4} & 89.2\se{2.1} & 80.8\se{2.3} & 86.2\se{2.4} & 77.9\se{1.9} & 91.2\se{1.4} & 81.5\se{1.7} & 88.9\se{0.9} & 80.7\se{0.9} \\
 & AdaST & \textbf{87.0}\se{3.3} & \textbf{80.2}\se{3.3} & \textbf{89.8}\se{1.4} & \textbf{81.5}\se{1.4} & \textbf{91.4}\se{1.6} & \textbf{83.9}\se{2.2} & \textbf{91.3}\se{1.8} & \textbf{83.1}\se{2.5} & \textbf{91.0}\se{1.5} & \textbf{84.0}\se{2.9} & \textbf{88.5}\se{2.3} & \textbf{80.5}\se{2.2} & \textbf{91.7}\se{1.3} & \textbf{82.2}\se{1.9} & \textbf{90.1}\se{0.8} & \textbf{82.2}\se{0.9} \\
\bottomrule
\end{tabular}}
\end{table}

\subsection{Comparisons Across Baseline}
\label{app:baselines}

Table~\ref{tab:baseline-full} gives the per-model, per-dataset breakdown behind
Figure~\ref{fig:baseline-comparison}.

\begin{table}[H]
\centering
\caption{Matched-budget hallucination detection at $n_T=25$, by model and
target dataset. Subscripts are bootstrap standard errors }
\label{tab:baseline-full}
\setlength{\tabcolsep}{2.0pt}
\resizebox{\textwidth}{!}{%
\begin{tabular}{ll*{7}{cc}}
\toprule
& & \multicolumn{2}{c}{NQ} & \multicolumn{2}{c}{TriviaQA} & \multicolumn{2}{c}{CoQA} & \multicolumn{2}{c}{HaluEval} & \multicolumn{2}{c}{SimpleQA} & \multicolumn{2}{c}{HotpotQA} & \multicolumn{2}{c}{Avg.} \\
\cmidrule(lr){3-4} \cmidrule(lr){5-6} \cmidrule(lr){7-8} \cmidrule(lr){9-10} \cmidrule(lr){11-12} \cmidrule(lr){13-14} \cmidrule(lr){15-16}
Model & Method & AUC & F1 & AUC & F1 & AUC & F1 & AUC & F1 & AUC & F1 & AUC & F1 & AUC & F1 \\
\midrule
OPT-6.7B & HaloScope & 57.3\se{1.5} & 52.9\se{0.6} & 61.3\se{2.7} & 55.6\se{1.9} & 53.6\se{1.6} & 51.2\se{1.6} & 68.0\se{3.6} & 61.9\se{3.1} & 61.4\se{3.0} & 52.9\se{1.8} & 51.7\se{2.6} & 49.7\se{0.8} & 58.9\se{2.2} & 54.0\se{1.6} \\
 & HARP & 66.0\se{1.2} & 52.4\se{1.4} & 67.7\se{1.0} & 54.5\se{2.6} & 63.8\se{0.7} & 55.2\se{1.8} & 81.8\se{0.5} & 68.3\se{3.7} & 60.8\se{3.2} & 49.7\se{2.6} & 59.9\se{2.2} & 49.1\se{1.8} & 66.6\se{3.0} & 54.9\se{2.6} \\
 & SinkProbe & 75.0\se{1.1} & 49.0\se{0.5} & 74.8\se{0.6} & 54.8\se{3.0} & 58.9\se{0.6} & 41.5\se{2.4} & 82.6\se{0.6} & 69.9\se{1.1} & 74.1\se{2.9} & 52.1\se{1.8} & \textbf{75.7}\se{1.0} & 52.7\se{2.5} & 73.5\se{2.8} & 53.3\se{3.4} \\
 & Target-only & 71.9\se{1.0} & 53.5\se{1.0} & 75.7\se{0.6} & \textbf{65.2}\se{0.6} & 65.8\se{1.0} & 60.7\se{1.2} & 84.6\se{0.3} & \textbf{77.8}\se{0.2} & 73.3\se{2.3} & 52.5\se{1.5} & 73.5\se{1.5} & \textbf{58.6}\se{1.7} & 74.1\se{2.3} & 61.4\se{3.5} \\
 & AdaST (Ours) & \textbf{82.2}\se{0.4} & \textbf{60.7}\se{0.7} & \textbf{79.9}\se{0.5} & 64.6\se{1.4} & \textbf{68.9}\se{0.4} & \textbf{62.5}\se{0.8} & \textbf{85.4}\se{0.2} & 77.3\se{0.8} & \textbf{79.9}\se{2.0} & \textbf{55.1}\se{1.3} & 75.2\se{1.3} & 57.3\se{1.3} & \underline{\textbf{78.6}}\se{2.2} & \underline{\textbf{62.9}}\se{2.9} \\
\midrule
Llama-3.1-8B & HaloScope & 64.9\se{3.0} & 56.0\se{2.0} & 70.8\se{4.1} & 64.7\se{3.4} & 54.5\se{1.8} & 51.7\se{2.0} & 77.6\se{2.8} & 66.5\se{3.6} & 57.3\se{1.9} & 49.3\se{1.7} & 56.2\se{3.3} & 50.0\se{3.3} & 63.5\se{3.5} & 56.4\se{2.8} \\
 & HARP & 69.3\se{1.3} & 59.1\se{1.6} & 80.9\se{0.5} & 69.2\se{2.4} & 62.7\se{0.9} & 53.8\se{2.4} & 82.7\se{0.5} & 69.8\se{2.4} & 58.8\se{1.6} & 46.1\se{1.4} & 65.2\se{2.2} & 49.8\se{2.9} & 69.9\se{3.6} & 58.0\se{3.7} \\
 & SinkProbe & 84.2\se{1.1} & 54.0\se{2.7} & 84.8\se{0.4} & 74.6\se{1.5} & 69.6\se{1.1} & 54.5\se{4.5} & 82.7\se{0.6} & 64.2\se{3.8} & 72.6\se{1.7} & 50.4\se{1.3} & 75.1\se{0.7} & 51.8\se{1.7} & 78.2\se{2.4} & 58.3\se{3.5} \\
 & Target-only & 74.0\se{2.9} & 61.2\se{1.1} & 82.7\se{0.3} & 76.5\se{0.5} & 74.3\se{0.8} & \textbf{68.1}\se{0.5} & 81.6\se{0.7} & 75.8\se{0.8} & 64.6\se{0.9} & 53.9\se{0.8} & 67.5\se{1.5} & 55.5\se{1.4} & 74.1\se{2.7} & 65.2\se{3.7} \\
 & AdaST (Ours) & \textbf{88.7}\se{0.2} & \textbf{71.1}\se{0.8} & \textbf{86.1}\se{0.2} & \textbf{78.9}\se{0.4} & \textbf{75.1}\se{0.6} & 67.8\se{1.0} & \textbf{84.8}\se{0.3} & \textbf{77.9}\se{0.7} & \textbf{81.3}\se{0.2} & \textbf{59.5}\se{0.9} & \textbf{79.6}\se{0.4} & \textbf{63.1}\se{0.5} & \underline{\textbf{82.6}}\se{1.8} & \underline{\textbf{69.7}}\se{2.9} \\
\midrule
Gemma-7B & HaloScope & 57.0\se{1.9} & 50.2\se{1.1} & 72.9\se{4.3} & 66.4\se{3.7} & 52.0\se{2.1} & 47.3\se{2.8} & 61.0\se{3.2} & 55.1\se{3.4} & 63.0\se{1.9} & 52.5\se{1.1} & 62.0\se{5.3} & 52.2\se{1.7} & 61.3\se{2.6} & 54.0\se{2.4} \\
 & HARP & 51.8\se{1.2} & 49.9\se{1.3} & 74.1\se{1.0} & 61.6\se{3.8} & 60.0\se{0.7} & 53.0\se{1.6} & 79.0\se{0.8} & 62.0\se{3.0} & 59.6\se{2.1} & 47.3\se{2.1} & 66.1\se{1.1} & 49.9\se{3.2} & 65.1\se{3.7} & 53.9\se{2.4} \\
 & SinkProbe & 74.7\se{2.6} & 48.7\se{0.0} & 84.8\se{0.6} & 65.9\se{4.2} & 63.9\se{1.2} & 47.5\se{5.2} & 82.5\se{1.1} & 69.5\se{3.2} & 70.5\se{1.5} & 50.8\se{1.0} & 74.9\se{2.5} & 56.8\se{2.7} & 75.2\se{2.9} & 56.5\se{3.4} \\
 & Target-only & 70.1\se{2.8} & 55.8\se{1.4} & 83.6\se{0.3} & 75.8\se{0.7} & 68.3\se{1.2} & 62.8\se{1.1} & 84.0\se{0.3} & 79.6\se{0.3} & 70.0\se{1.9} & 54.5\se{0.7} & 73.1\se{2.3} & 58.2\se{1.9} & 74.9\se{2.7} & 64.4\se{4.0} \\
 & AdaST (Ours) & \textbf{83.5}\se{0.3} & \textbf{60.8}\se{1.8} & \textbf{86.2}\se{0.2} & \textbf{76.9}\se{0.6} & \textbf{70.7}\se{0.6} & \textbf{62.9}\se{0.4} & \textbf{85.8}\se{0.2} & \textbf{80.7}\se{0.2} & \textbf{80.1}\se{0.5} & \textbf{55.8}\se{1.0} & \textbf{82.2}\se{0.9} & \textbf{61.6}\se{0.8} & \underline{\textbf{81.4}}\se{2.1} & \underline{\textbf{66.5}}\se{3.7} \\
\midrule
Mistral-7B & HaloScope & 47.4\se{1.3} & 47.4\se{1.6} & 53.3\se{1.5} & 49.3\se{1.8} & 51.0\se{6.7} & 50.7\se{2.3} & 51.3\se{2.4} & 45.5\se{2.0} & 54.9\se{1.2} & 49.8\se{0.6} & 55.2\se{3.1} & 49.9\se{1.2} & 52.2\se{1.1} & 48.8\se{0.7} \\
 & HARP & 69.5\se{0.9} & 52.0\se{2.2} & 69.7\se{0.7} & 57.4\se{1.2} & 73.8\se{1.2} & 55.9\se{1.1} & 65.0\se{0.9} & 50.5\se{2.9} & 68.8\se{0.7} & 52.0\se{3.2} & 70.4\se{0.7} & 50.5\se{1.7} & 69.5\se{1.1} & 53.1\se{1.1} \\
 & SinkProbe & \textbf{84.2}\se{0.4} & 55.9\se{3.0} & \textbf{84.5}\se{0.5} & 60.9\se{3.4} & 77.5\se{1.4} & 53.1\se{2.0} & 80.6\se{0.6} & 53.6\se{2.2} & 83.8\se{0.5} & 55.6\se{2.0} & \textbf{84.8}\se{0.8} & 52.2\se{2.5} & 82.5\se{1.1} & 55.2\se{1.2} \\
 & Target-only & 70.2\se{0.7} & 59.0\se{0.5} & 76.6\se{0.6} & 63.5\se{0.4} & 75.1\se{2.0} & 58.8\se{1.4} & 70.3\se{1.3} & 61.0\se{0.7} & 70.2\se{0.9} & 56.7\se{0.7} & 71.8\se{1.1} & 57.0\se{1.4} & 72.4\se{1.1} & 59.3\se{1.0} \\
 & AdaST (Ours) & 80.4\se{0.1} & \textbf{65.8}\se{0.4} & 83.9\se{0.2} & \textbf{69.9}\se{0.5} & \textbf{83.4}\se{0.5} & \textbf{62.1}\se{0.8} & \textbf{82.0}\se{0.2} & \textbf{67.8}\se{0.2} & \textbf{84.2}\se{0.1} & \textbf{66.6}\se{0.4} & 83.8\se{0.2} & \textbf{66.0}\se{0.3} & \underline{\textbf{83.0}}\se{0.5} & \underline{\textbf{66.3}}\se{1.0} \\
\midrule
DeepSeek-7B & HaloScope & 61.3\se{6.0} & 51.7\se{2.3} & 70.7\se{2.1} & 64.1\se{1.7} & 56.5\se{1.4} & 53.2\se{1.0} & 66.9\se{2.4} & 61.7\se{2.7} & 59.3\se{2.5} & 48.5\se{2.1} & 58.0\se{1.6} & 49.1\se{1.1} & 62.1\se{2.0} & 54.7\se{2.5} \\
 & HARP & 66.2\se{4.2} & 48.5\se{3.7} & 70.3\se{1.0} & 54.1\se{4.1} & 58.5\se{1.6} & 46.0\se{2.4} & 78.8\se{0.7} & 59.7\se{3.0} & 65.3\se{0.9} & 46.3\se{2.4} & 63.9\se{1.8} & 49.5\se{2.9} & 67.2\se{2.6} & 50.7\se{2.0} \\
 & SinkProbe & 82.2\se{1.0} & 52.3\se{2.2} & 77.1\se{0.9} & 56.6\se{3.6} & 64.4\se{1.0} & 47.1\se{5.0} & 80.9\se{0.8} & 68.4\se{3.1} & 66.1\se{1.6} & 49.7\se{1.0} & 69.8\se{1.6} & 51.3\se{2.0} & 73.4\se{2.9} & 54.2\se{2.8} \\
 & Target-only & 79.3\se{1.7} & 59.9\se{1.7} & 78.7\se{0.7} & 71.8\se{0.7} & 69.3\se{0.7} & 63.3\se{0.6} & 83.3\se{0.3} & 76.0\se{1.0} & 71.9\se{1.8} & 55.3\se{1.1} & 71.9\se{0.8} & 56.8\se{1.0} & 75.7\se{2.0} & 63.9\se{3.1} \\
 & AdaST (Ours) & \textbf{86.5}\se{0.7} & \textbf{63.7}\se{1.9} & \textbf{82.7}\se{0.3} & \textbf{72.7}\se{1.1} & \textbf{71.2}\se{0.6} & \textbf{64.7}\se{0.6} & \textbf{85.2}\se{0.1} & \textbf{78.5}\se{0.3} & \textbf{77.2}\se{0.6} & \textbf{55.9}\se{1.2} & \textbf{74.9}\se{0.4} & \textbf{57.9}\se{1.2} & \underline{\textbf{79.6}}\se{2.3} & \underline{\textbf{65.6}}\se{3.2} \\
\bottomrule
\end{tabular}}
\end{table}

\section{Additional Implementation Details}

\subsection{Layer Selection}
\label{app:layer}

For each model and target domain $T$, we select the representation layer using
labeled source data only. For each source domain $s\in[S]$, we divide its questions (or reasoning prefixes) into a $70\%$ fitting partition and a
$30\%$ validation partition. At each candidate layer $j$, we estimate a mean-difference direction between the positive and negative classes using the
fitting partition and evaluate its AUROC on the corresponding validation partition. Here, the positive class corresponds to truthful responses for
hallucination detection and correct next steps for reasoning-step correctness.
We select
\[
L_{m,T}
=
\arg\max_j
\frac{1}{S}
\sum_{s\in[S]}
\operatorname{AUROC}^{\mathrm{val}}_{m,s,j},
\]
where $m$ indexes the evaluated language model. The selected layer is then
fixed for all target-label budgets, cross-fitting folds, and experimental
repetitions for that model--target pair.

For hallucination detection, the five remaining QA datasets serve as source
domains when one dataset is held out as the target. For reasoning-step
correctness, the six remaining MATH subjects serve as source domains. Neither
labeled target data nor the target test set is used for layer selection,
preserving the target-label budget and avoiding target-test leakage. After
selecting $L_{m,T}$, the source detectors are refitted using all available
labeled data from their respective source domains.

\subsection{Intervention Direction and Activation Steering}

\label{app:intervention}

For detector $s\in[S]\cup\{T\}$, recall that its raw residual-space score is $\mathbf v_s^\top \mathbf h$. We denote it by $g_s(\mathbf h)$.

To place the scores of different detectors on a common scale, we standardize them using the labeled target responses. Specifically, we define the mean raw score of detector $s$ on the target set as
\[
\bar g_s
=
\frac{1}{n_T P R}
\sum_{q\in \mathcal Q_T}\sum_{p=1}^{P}\sum_{r=1}^{R}
\mathbf v_s^\top\mathbf h_{qpr},
\]
and define
\[
\kappa_s
=
\sqrt{
\frac{1}{n_T P R-1}
\sum_{q\in \mathcal Q_T}\sum_{p=1}^{P}\sum_{r=1}^{R}
\left(
\mathbf v_s^\top\mathbf h_{qpr}-\bar g_s
\right)^2
}.
\]
Thus, $\kappa_s$ is the empirical standard deviation of detector $s$'s raw scores on the labeled target responses. The standardized detector score is therefore
\[
\psi_s(\mathbf h)
=
\frac{\mathbf v_s^\top\mathbf h-\bar g_s}{\kappa_s}.
\]
Since $\bar g_s$ is constant with respect to $\mathbf h$, the direction in residual space along which the standardized detector score increases is
\[
\nabla_{\mathbf h}\psi_s(\mathbf h)
=
\frac{\mathbf v_s}{\kappa_s}.
\]

The final AdaST detector combines the standardized detector scores as
\[
\widehat f(\mathbf h)
=
\widehat b
+
\sum_{s\in[S]\cup\{T\}}
\widehat w_s
\psi_s(\mathbf h).
\]
Its corresponding residual-space direction is therefore
\[
\nabla_{\mathbf h}\widehat f(\mathbf h)
=
\sum_{s\in[S]\cup\{T\}}
\widehat w_s\frac{\mathbf v_s}{\kappa_s}.
\]
We normalize this vector to unit norm to obtain the steering direction $\mathbf u_{\mathrm{mix}}$ used in the intervention experiment. We next describe how the resulting direction is applied during generation. We implement activation steering using a forward hook on transformer block $L$. During the initial prompt pass, the hook adds $\alpha\mathbf{u}_{\mathrm{mix}}$ only to the block output at the final prompt position. During cached autoregressive decoding, the same update is applied to the current token position at every decoding step:
\[
\mathbf{h}_{t}^{(L)}
\leftarrow
\mathbf{h}_{t}^{(L)}
+
\alpha\mathbf{u}_{\mathrm{mix}}.
\]
The modified state is then passed to block $L+1$ to predict the next token. The steering direction $\mathbf{u}_{\mathrm{mix}}$ and strength $\alpha$ are fixed throughout the generation of a response. No earlier prompt positions or other transformer layers are modified.

For a unit steering direction
$\mathbf u_{\mathrm{mix}}$, we first compute its activation scale
\[
A_{\mathrm{mix}}
=
\max_{i\in\mathcal V}
\left|
\left\langle
\mathbf h_{i,-1}^{(L)},\mathbf u_{\mathrm{mix}}
\right\rangle
\right|,
\]
where $\mathcal V$ denotes the steering validation set and $\mathbf h_{i,-1}^{(L)}$ is the residual state of the final prompt token at layer $L$. We evaluate strengths of the form $\alpha=cA_{\mathrm{mix}}$ for $c\in\{0,0.1,0.3,1,3,10,20,40\}$ and select the value that maximizes mean BLEURT on the validation responses. Ties are resolved in favor of the smaller magnitude. The selected $\alpha$ is then frozen and evaluated on 500 disjoint test questions.

\section{Additional Lemmas}

\begin{lemma}[Sandwiched sample second moments.]
\label{lemma:sandwich}
Let $x_1,\dots,x_n\in\mathbb R^D$ be i.i.d. with
$\mathbb E\|x_\ell\|_2^2<\infty$ and
$\mathbb E[x_\ell x_\ell^\top]=\Sigma$, and let
$\widehat\Sigma=\frac1n\sum_{\ell=1}^{n}x_\ell x_\ell^\top$.
Let $A\in\mathbb R^{D\times m_A}$ and
$B\in\mathbb R^{D\times m_B}$ be fixed matrices whose numbers
of columns do not depend on $n$. With $D$, $A$, and $B$ fixed,
as $n\to\infty$,
\[
A^\top\widehat\Sigma B
\xrightarrow{\mathrm{a.s.}}
A^\top\Sigma B.
\]
If in addition $A^\top\Sigma A$ is positive definite, then
almost surely $A^\top\widehat\Sigma A$ is invertible for all
large $n$ and
$(A^\top\widehat\Sigma A)^{-1}\to(A^\top\Sigma A)^{-1}$.
\end{lemma}

\paragraph{Proof.}
Fix an entry $(a,b)$ with $a\le m_A$ and $b\le m_B$, and
let $e_a,e_b$ denote standard basis vectors of
$\mathbb R^{m_A}$ and $\mathbb R^{m_B}$. The entry is a
sample mean,
\[
\big(A^\top\widehat\Sigma B\big)_{ab}
=\frac1n\sum_{\ell=1}^{n}Y_\ell,
\qquad
Y_\ell
=\big(e_a^\top A^\top x_\ell\big)
 \big(e_b^\top B^\top x_\ell\big).
\]
The variables $Y_\ell$ are i.i.d. across $\ell$, being a
fixed function of the i.i.d. vectors $x_\ell$. They are
integrable by the Cauchy Schwarz inequality,
\[
\mathbb E|Y_\ell|
\le
\big(\mathbb E(e_a^\top A^\top x_\ell)^2\big)^{1/2}
\big(\mathbb E(e_b^\top B^\top x_\ell)^2\big)^{1/2}
=
\big(e_a^\top A^\top\Sigma Ae_a\big)^{1/2}
\big(e_b^\top B^\top\Sigma Be_b\big)^{1/2}
<\infty.
\]
Their mean is
$\mathbb E Y_\ell=(A^\top\Sigma B)_{ab}$.
Kolmogorov's strong law of large numbers gives
$\frac1n\sum_\ell Y_\ell\to(A^\top\Sigma B)_{ab}$ almost
surely. This holds on an event $\Omega_{ab}$ with
$\mathbb P(\Omega_{ab})=1$. There are $m_Am_B$ entries,
a finite number that does not depend on $n$, so
$\mathbb P(\bigcap_{a,b}\Omega_{ab})=1$.
On this event every entry converges. Entrywise convergence
of matrices of fixed size is convergence in every matrix
norm, which proves the first claim.

For the second claim, work on the same probability one
event with $A=B$. By Weyl's inequality,
\[
\lambda_{\min}(A^\top\widehat\Sigma A)
\ge
\lambda_{\min}(A^\top\Sigma A)
-
\|A^\top\widehat\Sigma A-A^\top\Sigma A\|_{\mathrm{op}}.
\]
The right side converges to
$\lambda_{\min}(A^\top\Sigma A)>0$, so
$A^\top\widehat\Sigma A$ is invertible for all large $n$.
Continuity of matrix inversion gives
$(A^\top\widehat\Sigma A)^{-1}\to(A^\top\Sigma A)^{-1}$.

\begin{lemma}[Projected noise sample means.]
\label{lemma:projected}
Let $(x_\ell,\varepsilon_\ell)$, $\ell=1,\dots,n$, be
i.i.d. pairs with
$\mathbb E[\varepsilon_\ell\mid x_\ell]=0$,
$\mathbb E\varepsilon_\ell^2<\infty$, and
$\mathbb E\|x_\ell\|_2^2<\infty$.
Write $\Sigma=\mathbb E[x_\ell x_\ell^\top]$,
$X=[x_1,\dots,x_n]\in\mathbb R^{D\times n}$, and
$\varepsilon=(\varepsilon_1,\dots,\varepsilon_n)^\top$.
Let $A\in\mathbb R^{D\times m_A}$ be a fixed matrix whose
number of columns does not depend on $n$.
With $D$ and $A$ fixed, as $n\to\infty$,
\[
\frac1nA^\top X\varepsilon
=
\frac1n\sum_{\ell=1}^{n}(A^\top x_\ell)\varepsilon_\ell
\xrightarrow{\mathrm{a.s.}}0.
\]
\end{lemma}

\paragraph{Proof.}
Fix a coordinate $a\le m_A$ and let $e_a$ be the $a$th
standard basis vector of $\mathbb R^{m_A}$.
The coordinate is the sample mean
$\frac1n\sum_\ell Y_\ell$ with
$Y_\ell=(e_a^\top A^\top x_\ell)\varepsilon_\ell$.
These variables are i.i.d. across $\ell$ and integrable
by the Cauchy Schwarz inequality,
\[
\mathbb E|Y_\ell|
\le
\big(\mathbb E(e_a^\top A^\top x_\ell)^2\big)^{1/2}
\big(\mathbb E\varepsilon_\ell^2\big)^{1/2}
=
\big(e_a^\top A^\top\Sigma Ae_a\big)^{1/2}
\big(\mathbb E\varepsilon_\ell^2\big)^{1/2}
<\infty.
\]
Their mean is zero by the tower property,
\[
\mathbb E Y_\ell
=
\mathbb E\Big[
(e_a^\top A^\top x_\ell)
\mathbb E[\varepsilon_\ell\mid x_\ell]
\Big]
=0.
\]
Kolmogorov's strong law of large numbers gives
$\frac1n\sum_\ell Y_\ell\to0$ almost surely.
Since there are finitely many coordinates, the entire vector
converges to zero almost surely.

\section{Proof of Theorem~\ref{thrm:risk}}
\label{app:proof-risk}

\subsection{Model and Assumptions}
\label{app:risk-assumptions}

We use the Gaussian linear model and fixed prediction
subspaces defined in the main text. Observations are i.i.d.
within each domain, and datasets from different domains
are independent. Noise is independent of features, with
mean zero and fixed finite variance. The target test
observation is independent of all training data.

All source and target sample sizes tend to infinity, while
$S$, $D$, $k$, the prediction subspaces, and the population
distributions remain fixed. Assume that the projected
covariances $\mU_s^\top\mSigma_s\mU_s$ are positive definite
in every domain and that
$\overline{\mH}=\mQ^\top\mSigma_T\mQ\succ0$.
The target fold assignments are fixed across the asymptotic sequence and independent of the observations, with $n_{T,f}/n_T\to\rho_f\in(0,1)$ for $f=1,2$.

\subsection{Two-Fold Out-of-Fold Weight Estimation}

Write $\mX_s=[\vx_{s,1},\ldots,\vx_{s,n_s}]$,
$\vy_s=(y_{s,1},\ldots,y_{s,n_s})^\top$, and
$\vvarepsilon_s=(\varepsilon_{s,1},\ldots,\varepsilon_{s,n_s})^\top$.
For each source domain, fit the projected linear regression
\[
\widehat\vbeta_s
=
\arg\min_{\vbeta\in\mathbb R^k}
\frac1{n_s}
\|\vy_s-\mX_s^\top\mU_s\vbeta\|_2^2,
\qquad
\widehat\vtheta_s=\mU_s\widehat\vbeta_s.
\]
Split the target observations into two disjoint folds with
sample sizes $n_{T,1}+n_{T,2}=n_T$.
Using only fold $f\in\{1,2\}$, fit the projected linear
regression in the target subspace
\[
\widehat\vbeta_{T,f}
=
\arg\min_{\vbeta\in\mathbb R^k}
\frac1{n_{T,f}}
\|\vy_{T,f}-\mX_{T,f}^\top\mU_T\vbeta\|_2^2,
\qquad
\widehat\vtheta_{T,f}=\mU_T\widehat\vbeta_{T,f}.
\]
The aggregation weights
$\vw=(w_1,\ldots,w_S,w_{S+1})^\top$ are estimated by
\begin{align*}
\widehat\vw
=
\arg\min_{\substack{\vw\ge0\\\mathbf1^\top\vw=1}}
\frac1{n_T}\Bigg[&
\left\|
\vy_{T,1}
-\mX_{T,1}^\top
\left(
\sum_{s=1}^S w_s\widehat\vtheta_s
+w_{S+1}\widehat\vtheta_{T,2}
\right)
\right\|_2^2\\
&+
\left\|
\vy_{T,2}
-\mX_{T,2}^\top
\left(
\sum_{s=1}^S w_s\widehat\vtheta_s
+w_{S+1}\widehat\vtheta_{T,1}
\right)
\right\|_2^2
\Bigg].
\end{align*}
Any measurable minimizer can be used if the empirical
optimum is not unique.

\subsection{Risk Calculation}

The projected linear regression using the full target
dataset is
\[
\widehat\vbeta_T
=
\arg\min_{\vbeta\in\mathbb R^k}
\frac1{n_T}
\|\vy_T-\mX_T^\top\mU_T\vbeta\|_2^2,
\qquad
\widehat\vtheta_{S+1}=\mU_T\widehat\vbeta_T.
\]
Hence, the transferred estimator is
\begin{equation}
\widehat{\vtheta}_{\text{AdaST}}
=
\sum_{s=1}^S\widehat w_s\widehat\vtheta_s
+\widehat w_{S+1}\widehat\vtheta_{S+1},
\label{eq:theta_decompose}
\end{equation}
with the last coordinate of the weight vector corresponds to the target domain and $\widehat w_{S+1}=\widehat w_T$. For an independent target observation $(\vx_0,y_0)$,
the noise has mean zero and is independent of the
features, so
\[
\mathcal R_T(\widehat{\vtheta}_{\text{AdaST}})
=
\mathbb E\big[
(y_0-\vx_0^\top\widehat{\vtheta}_{\text{AdaST}})^2
\mid\widehat{\vtheta}_{\text{AdaST}}
\big]
=
\sigma_T^2
+
(\widehat{\vtheta}_{\text{AdaST}}-\vtheta_T^\star)^\top
\mSigma_T
(\widehat{\vtheta}_{\text{AdaST}}-\vtheta_T^\star).
\]
Using~\eqref{eq:theta_decompose} and the constraint
$\mathbf1^\top\widehat\vw=1$,
\begin{align*}
\widehat{\vtheta}_{\text{AdaST}}-\vtheta_T^\star
&=
\sum_{s=1}^S\widehat w_s\widehat\vtheta_s
+\widehat w_{S+1}\widehat\vtheta_{S+1}
-
\left(
\sum_{s=1}^S\widehat w_s+\widehat w_{S+1}
\right)\vtheta_T^\star\\
&=
\sum_{s=1}^S
\widehat w_s(\widehat\vtheta_s-\vtheta_T^\star)
+
\widehat w_{S+1}
(\widehat\vtheta_{S+1}-\vtheta_T^\star).
\end{align*}
Here, $\widehat\vtheta_{S+1}$ is the target-only coefficient. For each source task, the source--target shift,
representation bias, and estimation error are
\[
\mDelta_s=\vtheta_s^\star-\vtheta_T^\star,
\qquad
\vb_s=\vtheta_s^\circ-\vtheta_s^\star,
\qquad
\vxi_s=\widehat\vtheta_s-\vtheta_s^\circ.
\]
The effective discrepancy between the projected source coefficient and the target coefficient is therefore
\[
\widetilde\mDelta_s
=
\vtheta_s^\circ-\vtheta_T^\star
=
\mDelta_s+\vb_s,
\qquad
\widehat\vtheta_s-\vtheta_T^\star
=
\widetilde\mDelta_s+\vxi_s.
\]
For the target task, the representation bias and
estimation error are
\[
\vb_T=\vtheta_T^\circ-\vtheta_T^\star,
\qquad
\vxi_{S+1}=\widehat\vtheta_{S+1}-\vtheta_T^\circ.
\]
Hence,
\[
\widehat{\vtheta}_{\text{AdaST}}-\vtheta_T^\star
=
\sum_{s=1}^S
\widehat w_s(\widetilde\mDelta_s+\vxi_s)
+
\widehat w_{S+1}(\vb_T+\vxi_{S+1}),
\]
and expanding the quadratic risk gives
\begin{align*}
\mathcal R_T(\widehat{\vtheta}_{\text{AdaST}})
&=
\sigma_T^2
+
\sum_{s=1}^S\sum_{s'=1}^S
\widehat w_s\widehat w_{s'}
\Big[
\widetilde\mDelta_s^\top\mSigma_T\widetilde\mDelta_{s'}
+
\widetilde\mDelta_s^\top\mSigma_T\vxi_{s'}
+
\vxi_s^\top\mSigma_T\widetilde\mDelta_{s'}
+
\vxi_s^\top\mSigma_T\vxi_{s'}
\Big]\\
&\quad+
2\sum_{s=1}^S
\widehat w_s\widehat w_{S+1}
\Big[
\widetilde\mDelta_s^\top\mSigma_T\vb_T
+
\widetilde\mDelta_s^\top\mSigma_T\vxi_{S+1}
+
\vxi_s^\top\mSigma_T\vb_T
+
\vxi_s^\top\mSigma_T\vxi_{S+1}
\Big]\\
&\quad+
\widehat w_{S+1}^2
\Big[
\vb_T^\top\mSigma_T\vb_T
+
\vb_T^\top\mSigma_T\vxi_{S+1}
+
\vxi_{S+1}^\top\mSigma_T\vb_T
+
\vxi_{S+1}^\top\mSigma_T\vxi_{S+1}
\Big].
\end{align*}

We first show that the estimation errors vanish.
Let $\widehat\mSigma_s=\mX_s\mX_s^\top/n_s$.
Substituting
$\vy_s=\mX_s^\top\vtheta_s^\star+\vvarepsilon_s$
into the projected least-squares normal equations gives
\begin{align*}
\widehat\vbeta_s
&=
(\mU_s^\top\widehat\mSigma_s\mU_s)^{-1}
\mU_s^\top\widehat\mSigma_s\vtheta_s^\star\\
&\quad+
(\mU_s^\top\widehat\mSigma_s\mU_s)^{-1}
\frac1{n_s}\mU_s^\top\mX_s\vvarepsilon_s.
\end{align*}
By Lemma~\ref{lemma:sandwich},
\[
\mU_s^\top\widehat\mSigma_s\mU_s
\xrightarrow{a.s.}
\mU_s^\top\mSigma_s\mU_s,
\qquad
\mU_s^\top\widehat\mSigma_s\vtheta_s^\star
\xrightarrow{a.s.}
\mU_s^\top\mSigma_s\vtheta_s^\star.
\]
The first limit is positive definite, so the sample
matrix is almost surely invertible for all sufficiently large sample sizes
and its inverse converges to
$(\mU_s^\top\mSigma_s\mU_s)^{-1}$.
By Lemma~\ref{lemma:projected},
\[
\frac1{n_s}\mU_s^\top\mX_s\vvarepsilon_s
\xrightarrow{a.s.}0.
\]
Therefore,
\[
\widehat\vbeta_s
\xrightarrow{a.s.}
(\mU_s^\top\mSigma_s\mU_s)^{-1}
\mU_s^\top\mSigma_s\vtheta_s^\star
=
\vbeta_s^\circ,
\]
and hence
\[
\vxi_s
=
\widehat\vtheta_s-\vtheta_s^\circ
=
\mU_s(\widehat\vbeta_s-\vbeta_s^\circ)
\xrightarrow{a.s.}0.
\]
The same argument with $s=T$ defines $\vbeta_T^\circ$
and applies to the two target-fold estimators and the
full-sample target estimator.
Defining
$\vxi_{T,f}=\widehat\vtheta_{T,f}-\vtheta_T^\circ$
for $f=1,2$, we obtain
\[
\vxi_{T,1}\xrightarrow{a.s.}0,
\qquad
\vxi_{T,2}\xrightarrow{a.s.}0,
\qquad
\vxi_{S+1}\xrightarrow{a.s.}0.
\]

\subsection{Asymptotic Out-of-Fold Weights}

For fold $f\in\{1,2\}$, substituting the target model and
using $\mathbf1^\top\vw=1$, the residual in the
out-of-fold objective becomes
\[
\vvarepsilon_{T,f}
-
\mX_{T,f}^\top
\left(
\sum_{s=1}^S
w_s(\widetilde\mDelta_s+\vxi_s)
+
w_{S+1}(\vb_T+\vxi_{T,3-f})
\right).
\]
Define
\[
\widehat\mE_f
=
\begin{bmatrix}
\widetilde\mDelta_1+\vxi_1&
\cdots&
\widetilde\mDelta_S+\vxi_S&
\vb_T+\vxi_{T,3-f}
\end{bmatrix}.
\]
Expanding the squared residuals, the out-of-fold
objective is
\[
L(\vw)
=
\frac1{n_T}\|\vvarepsilon_T\|_2^2
-
2\vw^\top\widehat\vg
+
\vw^\top\widehat\mH\vw,
\]
where
\[
\widehat\vg
=
\frac1{n_T}\sum_{f=1}^2
\widehat\mE_f^\top\mX_{T,f}\vvarepsilon_{T,f},
\qquad
\widehat\mH
=
\frac1{n_T}\sum_{f=1}^2
\widehat\mE_f^\top
\mX_{T,f}\mX_{T,f}^\top
\widehat\mE_f.
\]
For $s\le S$, the source predictor is the same on both
target folds, so
\begin{align*}
\widehat g_s
&=
\frac1{n_T}
(\widetilde\mDelta_s+\vxi_s)^\top
\mX_T\vvarepsilon_T\\
&=
\frac1{n_T}
\widetilde\mDelta_s^\top\mX_T\vvarepsilon_T
+
(\widehat\vbeta_s-\vbeta_s^\circ)^\top
\frac1{n_T}\mU_s^\top\mX_T\vvarepsilon_T
\xrightarrow{a.s.}0.
\end{align*}
Here Lemma~\ref{lemma:projected} applies with
$A=\widetilde\mDelta_s$ and $A=\mU_s$, and
$\widehat\vbeta_s-\vbeta_s^\circ\xrightarrow{a.s.}0$.
For the target coordinate, the same argument gives
\begin{align*}
\widehat g_{S+1}
&=
\frac1{n_T}\sum_{f=1}^2
(\vb_T+\vxi_{T,3-f})^\top
\mX_{T,f}\vvarepsilon_{T,f}\\
&=
\sum_{f=1}^2\frac{n_{T,f}}{n_T}
\left[
\frac1{n_{T,f}}\vb_T^\top\mX_{T,f}\vvarepsilon_{T,f}
+
(\widehat\vbeta_{T,3-f}-\vbeta_T^\circ)^\top
\frac1{n_{T,f}}\mU_T^\top\mX_{T,f}\vvarepsilon_{T,f}
\right]
\xrightarrow{a.s.}0.
\end{align*}
Thus $\widehat\vg\xrightarrow{a.s.}0$.

We next consider $\widehat\mH$. For $s,s'\le S$,
\begin{align*}
\widehat H_{ss'}
&=
(\widetilde\mDelta_s+\vxi_s)^\top
\widehat\mSigma_T
(\widetilde\mDelta_{s'}+\vxi_{s'})\\
&=
\widetilde\mDelta_s^\top
\widehat\mSigma_T\widetilde\mDelta_{s'}
+
\widetilde\mDelta_s^\top\widehat\mSigma_T\vxi_{s'}
+
\vxi_s^\top\widehat\mSigma_T\widetilde\mDelta_{s'}
+
\vxi_s^\top\widehat\mSigma_T\vxi_{s'},
\end{align*}
where $\widehat\mSigma_T=\mX_T\mX_T^\top/n_T$.
Applying Lemma~\ref{lemma:sandwich} to coordinate vectors
gives $\widehat\mSigma_T\xrightarrow{a.s.}\mSigma_T$ in
operator norm because $D$ is fixed.
Each term involving an estimation error vanishes.
For example,
\[
|\widetilde\mDelta_s^\top\widehat\mSigma_T\vxi_{s'}|
\le
\|\widetilde\mDelta_s\|_2
\|\widehat\mSigma_T\|_{\mathrm{op}}
\|\vxi_{s'}\|_2
\xrightarrow{a.s.}0,
\qquad
|\vxi_s^\top\widehat\mSigma_T\vxi_{s'}|
\le
\|\vxi_s\|_2
\|\widehat\mSigma_T\|_{\mathrm{op}}
\|\vxi_{s'}\|_2
\xrightarrow{a.s.}0.
\]
Consequently,
\[
\widehat H_{ss'}
\xrightarrow{a.s.}
\widetilde\mDelta_s^\top
\mSigma_T
\widetilde\mDelta_{s'}.
\]
For $s\le S$,
\[
\widehat H_{s,S+1}
=
\sum_{f=1}^2\frac{n_{T,f}}{n_T}
(\widetilde\mDelta_s+\vxi_s)^\top
\widehat\mSigma_{T,f}
(\vb_T+\vxi_{T,3-f}),
\]
where
$\widehat\mSigma_{T,f}
=\mX_{T,f}\mX_{T,f}^\top/n_{T,f}$.
For each fold, expanding this product gives
\begin{align*}
&(\widetilde\mDelta_s+\vxi_s)^\top
\widehat\mSigma_{T,f}
(\vb_T+\vxi_{T,3-f})\\
&\quad=
\widetilde\mDelta_s^\top\widehat\mSigma_{T,f}\vb_T
+
\widetilde\mDelta_s^\top
\widehat\mSigma_{T,f}\vxi_{T,3-f}
+
\vxi_s^\top\widehat\mSigma_{T,f}\vb_T
+
\vxi_s^\top\widehat\mSigma_{T,f}\vxi_{T,3-f}\\
&\quad\xrightarrow{a.s.}
\widetilde\mDelta_s^\top\mSigma_T\vb_T.
\end{align*}
The error terms vanish by the same norm bounds, since
$\widehat\mSigma_{T,f}\xrightarrow{a.s.}\mSigma_T$ and
$\vxi_{T,3-f}\xrightarrow{a.s.}0$.
Using $\sum_f n_{T,f}/n_T=1$ gives
\[
\widehat H_{s,S+1}
\xrightarrow{a.s.}
\widetilde\mDelta_s^\top\mSigma_T\vb_T.
\]
Finally,
\begin{align*}
\widehat H_{S+1,S+1}
&=
\sum_{f=1}^2\frac{n_{T,f}}{n_T}
(\vb_T+\vxi_{T,3-f})^\top
\widehat\mSigma_{T,f}
(\vb_T+\vxi_{T,3-f})\\
&=
\sum_{f=1}^2\frac{n_{T,f}}{n_T}
\Big[
\vb_T^\top\widehat\mSigma_{T,f}\vb_T
+
2\vb_T^\top\widehat\mSigma_{T,f}\vxi_{T,3-f}
+
\vxi_{T,3-f}^\top
\widehat\mSigma_{T,f}\vxi_{T,3-f}
\Big]\\
&\xrightarrow{a.s.}
\vb_T^\top\mSigma_T\vb_T.
\end{align*}
These entrywise limits give
\[
\widehat\mH
\xrightarrow{a.s.}
\begin{bmatrix}
\widetilde\mDelta_1&\cdots&\widetilde\mDelta_S&\vb_T
\end{bmatrix}^\top
\mSigma_T
\begin{bmatrix}
\widetilde\mDelta_1&\cdots&\widetilde\mDelta_S&\vb_T
\end{bmatrix}
=
\mQ^\top\mSigma_T\mQ
=
\overline{\mH}.
\]

The limiting weights minimize
$\vw^\top\overline{\mH}\vw$ subject to
$\vw\ge0$ and $\mathbf1^\top\vw=1$.
The feasible set is compact and the objective is strictly
convex because $\overline{\mH}\succ0$, so its minimizer
$\vw^\star$ exists and is unique.
Every feasible vector satisfies $\|\vw\|_2\le1$,
giving the uniform bound
\[
\sup_{\substack{\vw\ge0\\\mathbf1^\top\vw=1}}
\left|
L(\vw)
-\frac1{n_T}\|\vvarepsilon_T\|_2^2
-\vw^\top\overline{\mH}\vw
\right|
\le
2\|\widehat\vg\|_2
+
\|\widehat\mH-\overline{\mH}\|_{\mathrm{op}}
\xrightarrow{a.s.}0.
\]
To show that the empirical minimizer converges, take
any feasible $\vw$. The segment from $\vw^\star$ to
$\vw$ is feasible, and optimality at $\vw^\star$
makes the directional derivative nonnegative.
Thus
$(\vw-\vw^\star)^\top\overline{\mH}\vw^\star\ge0$,
and
\begin{align*}
\vw^\top\overline{\mH}\vw
-
(\vw^\star)^\top\overline{\mH}\vw^\star
&=
2(\vw-\vw^\star)^\top\overline{\mH}\vw^\star
+
(\vw-\vw^\star)^\top
\overline{\mH}
(\vw-\vw^\star)\\
&\ge
\lambda_{\min}(\overline{\mH})
\|\vw-\vw^\star\|_2^2.
\end{align*}
Since $L(\widehat\vw)\le L(\vw^\star)$, applying the
uniform bound to both weights gives
\begin{align*}
\lambda_{\min}(\overline{\mH})
\|\widehat\vw-\vw^\star\|_2^2
&\le
\widehat\vw^\top\overline{\mH}\widehat\vw
-
(\vw^\star)^\top\overline{\mH}\vw^\star\\
&\le
4\|\widehat\vg\|_2
+
2\|\widehat\mH-\overline{\mH}\|_{\mathrm{op}}
\xrightarrow{a.s.}0.
\end{align*}
Because $\lambda_{\min}(\overline{\mH})>0$, this proves
\[
\widehat\vw
\xrightarrow{a.s.}
\vw^\star
=
\arg\min_{\substack{\vw\ge0\\\mathbf1^\top\vw=1}}
\vw^\top\overline{\mH}\vw.
\]

Returning to the risk expansion, every weight lies in
$[0,1]$. Terms involving estimation errors therefore
vanish. For example,
\[
|\widehat w_s\widehat w_{s'}
\widetilde\mDelta_s^\top\mSigma_T\vxi_{s'}|
\le
\|\widetilde\mDelta_s\|_2
\|\mSigma_T\|_{\mathrm{op}}
\|\vxi_{s'}\|_2
\xrightarrow{a.s.}0,
\]
and terms involving two errors are bounded by
$\|\mSigma_T\|_{\mathrm{op}}$ times the product of
their norms. Since there are finitely many terms,
substituting the weight limits yields
\begin{align*}
\mathcal R_T(\widehat{\vtheta}_{\text{AdaST}})
&\xrightarrow{a.s.}
\sigma_T^2
+
\sum_{s=1}^S\sum_{s'=1}^S
w_s^\star w_{s'}^\star
\widetilde\mDelta_s^\top
\mSigma_T
\widetilde\mDelta_{s'}\\
&\quad+
2\sum_{s=1}^S
w_s^\star w_{S+1}^\star
\widetilde\mDelta_s^\top\mSigma_T\vb_T
+
(w_{S+1}^\star)^2\vb_T^\top\mSigma_T\vb_T\\
&=
\sigma_T^2
+
(\vw^\star)^\top\overline{\mH}\vw^\star.
\end{align*}
The target-only estimator has limiting risk
$\sigma_T^2+\vb_T^\top\mSigma_T\vb_T
=\sigma_T^2+\overline H_{S+1,S+1}$.
Since $\mathbf e_{S+1}$ is feasible, optimality gives
\[
(\vw^\star)^\top\overline{\mH}\vw^\star
\le
\mathbf e_{S+1}^\top\overline{\mH}\mathbf e_{S+1}
=
\overline H_{S+1,S+1}.
\]
Equality holds exactly when
$\vw^\star=\mathbf e_{S+1}$, since the limiting
minimizer is unique. This completes the proof.

\section{Finite-Sample Decay of the Target Weight}
\label{app:finite-rates}

We quantify the \textbf{finite-sample} error under the Gaussian-design and fixed-subspace assumptions of Theorem~\ref{thrm:risk}. Write $\widehat w_T=\widehat w_{S+1}$,
$\lambda=\lambda_{\min}(\overline{\mH})>0$, and
\[
r_n^2
=
\sum_{s=1}^S\frac{1}{n_s-k-1}
+
\sum_{f=1}^2\frac{1}{n_{T,f}-k-1}.
\]
Assume all source and target-fold sample sizes exceed $k+1$. Below, $C$ denotes a constant depending only on the fixed population parameters and dimensions.

For source domain $s$, let
$\vxi_s=\widehat\vtheta_s-\vtheta_s^\circ$.
The projected residual has variance
$\nu_s^2=\sigma_s^2+\vb_s^\top\mSigma_s\vb_s$
and is independent of the projected covariates under
Gaussian design. The Gaussian inverse-Gram moment gives
\begin{equation}
\mathbb E\|\vxi_s\|_2^2
=
\frac{\nu_s^2}{n_s-k-1}
\operatorname{tr}\!\left[
(\mU_s^\top\mSigma_s\mU_s)^{-1}
\right].
\label{eq:finite-fit-error}
\end{equation}
The same identity applies to target-fold estimation errors,
with $s=T$ and sample size $n_{T,f}$.

Next, use the out-of-fold matrices $\widehat\mH$, $\widehat\vg$, and $\widehat\mE_f$ defined in Appendix~\ref{app:proof-risk}. Set
\[
p_f=n_{T,f}/n_T,\qquad
\bm{\Xi}_f=\widehat\mE_f-\mQ,\qquad
\widehat\mSigma_{T,f}=\frac{\mX_{T,f}\mX_{T,f}^\top}{n_{T,f}}.
\]
Since $\bm{\Xi}_f$ contains source fitting errors and the target fitting error from fold $3-f$, it is independent of the observations in fold $f$. Expanding $\mathbb E\|\bm{\Xi}_f\|_{\mathrm F}^2=\sum_{s=1}^S\mathbb E\|\vxi_s \|^2_2 + \mathbb E\|\vxi_{T,3-f} \|^2_2$ and with~\eqref{eq:finite-fit-error} gives $\mathbb E\|\bm{\Xi}_f\|_{\mathrm F}^2\le Cr_n^2$, where $C=\max_s \nu_s^2\operatorname{tr}\!\left[(\mU_s^\top\mSigma_s\mU_s)^{-1}\right]$. Expanding the quadratic gives
\begin{align*}
\widehat\mH-\overline{\mH}
&=
\mQ^\top(\widehat\mSigma_T-\mSigma_T)\mQ\\
&\quad+
\sum_{f=1}^2p_f
\left[
\mQ^\top\widehat\mSigma_{T,f}\bm{\Xi}_f
+\bm{\Xi}_f^\top\widehat\mSigma_{T,f}\mQ
+\bm{\Xi}_f^\top\widehat\mSigma_{T,f}\bm{\Xi}_f
\right].
\end{align*}
Since the sample covariance $\widehat{\mSigma}_T$ concentrates around its population counterpart $\mSigma_T$, Gaussian covariance moment bounds yield
$\mathbb E\left\|
\mQ^\top(\widehat{\mSigma}_T-\mSigma_T)\mQ
\right\|
\le C n_T^{-1/2}$. Since $\mathbb E\|\widehat\mSigma_{T,f}\|_{\mathrm{op}}\le\operatorname{tr}(\mSigma_T)$, then $\mathbb E\|\mQ^\top\widehat\mSigma_{T,f}\bm{\Xi}_f \|_{\mathrm{op}}\le \mathbb E\|\mQ\|_{\mathrm{op}}\operatorname{tr}(\mSigma_T)\sqrt{\mathbb E\|\bm{\Xi}_f\|_{\mathrm F}^2}=\mathcal{O}(r_n)$. Similarly, the quadratic term is bounded by $\mathcal{O}(r_n^2)$.

For the variance term, 
\[
\widehat\vg
=
\sum_{f=1}^2p_f(\mQ+\bm{\Xi}_f)^\top\frac{\mX_{T,f}\vvarepsilon_{T,f}}{n_{T,f}},
\qquad
\mathbb E\|\frac{\mX_{T,f}\vvarepsilon_{T,f}}{n_{T,f}}\|_2^2
=
\frac{\sigma_T^2\operatorname{tr}(\mSigma_T)}{n_{T,f}}.
\]
Similar calculations as for $\widehat\mH-\overline{\mH}$ give
$\mathbb E\|\widehat\vg\|_2
\le C(1+r_n)\sum_f p_f/\sqrt{n_{T,f}}$.
Together,
\begin{equation}
\mathbb E \left[\|\widehat\mH-\overline{\mH}\|_{\mathrm{op}}
+\|\widehat\vg\|_2 \right] \le C(r_n+r_n^2).
\label{eq:finite-objective-error}
\end{equation}

Finally, let
\[
\vd=\widehat{\vw}-\vw^\star.
\]
Since $\widehat{\vw}$ minimizes the empirical quadratic objective over the
simplex and $\vw^\star$ is feasible, the first-order optimality condition
gives
\[
(\vw^\star-\widehat{\vw})^\top
\left(
\widehat{\mH}\widehat{\vw}-\widehat{\vg}
\right)
\ge 0,
\]
or equivalently,
\[
\vd^\top
\left(
\widehat{\mH}\widehat{\vw}-\widehat{\vg}
\right)
\le 0.
\]
Similarly, since $\vw^\star$ minimizes the population quadratic objective
over the simplex and $\widehat{\vw}$ is feasible,
\[
(\widehat{\vw}-\vw^\star)^\top
\overline{\mH}\vw^\star
\ge 0,
\]
so that
\[
\vd^\top\overline{\mH}\vw^\star\ge0.
\]

Since $\lambda=\lambda_{\min}(\overline{\mH})$, we have
\begin{align*}
\lambda\|\vd\|_2^2
&\le
\vd^\top\overline{\mH}\vd\le
\vd^\top\overline{\mH}\widehat{\vw}\\
&=
\vd^\top
\left[
(\overline{\mH}-\widehat{\mH})\widehat{\vw}
+
\widehat{\mH}\widehat{\vw}
\right]\\
&=
\vd^\top
\left[
(\overline{\mH}-\widehat{\mH})\widehat{\vw}
+
\widehat{\vg}
\right]
+
\vd^\top
\left(
\widehat{\mH}\widehat{\vw}-\widehat{\vg}
\right)\\
&\le
\vd^\top
\left[
(\overline{\mH}-\widehat{\mH})\widehat{\vw}
+
\widehat{\vg}
\right]\\
&\le
\left(
\|\widehat{\mH}-\overline{\mH}\|_{\mathrm{op}}
+
\|\widehat{\vg}\|_2
\right)
\|\vd\|_2.
\end{align*}

Therefore
$\|\widehat\vw-\vw^\star\|_2\le (\|\widehat\mH-\overline{\mH}\|_{\mathrm{op}} +\|\widehat\vg\|_2)/\lambda$.
Taking expectations gives
\begin{equation}
\mathbb E\|\widehat\vw-\vw^\star\|_2
\le\frac{C}{\lambda}(r_n+r_n^2).
\label{eq:weight-mean-rate}
\end{equation}
In any setting whose population solution has $w_T^\star=0$,
simplex nonnegativity then gives
\begin{equation}
0\le\mathbb E\widehat w_T
\le\frac{C}{\lambda}(r_n+r_n^2).
\label{eq:target-weight-decay}
\end{equation}
In the empirically-interested regime where source samples are more abundant than target samples, with $n_s\ge c n_T$ for a fixed $c\ge1$ and $p_f\to\rho_f\in(0,1)$ for $f=1,2$, we have $r_n=O(n_T^{-1/2})$. Consequently, if $w_T^\star=0$,
\[
\mathbb E\widehat w_T=O(n_T^{-1/2}).
\]
This establishes a vanishing upper bound on the expected target weight. More generally, the same argument gives $\mathbb E|\widehat w_T-w_T^\star|=O(n_T^{-1/2})$, including when $w_T^\star>0$.


\end{document}